\documentclass[11pt]{article}

\usepackage[preprint]{acl}
\usepackage{amsmath}
\usepackage{booktabs}  
\usepackage{multirow}  
\usepackage{graphicx}  
\usepackage{comment}
\usepackage{adjustbox}
\usepackage{times}
\usepackage{latexsym}

\usepackage[T1]{fontenc}

\usepackage[utf8]{inputenc}

\usepackage{microtype}

\usepackage{inconsolata}

\usepackage{graphicx}

\usepackage{makecell}
\usepackage{nicefrac}
\usepackage{amssymb}
\usepackage{svg}
\graphicspath{{figures/}{../outputs/}}

\usepackage{booktabs}
\usepackage{array}
\usepackage{adjustbox}
\usepackage[table]{xcolor}
\usepackage{pifont}

\newcommand{\cmark}{\textcolor{green!60!black}{\ding{51}}}
\newcommand{\xmark}{\textcolor{red!70!black}{\ding{55}}}

\usepackage{listings}
\usepackage{xcolor}

\lstdefinestyle{promptstyle}{
    basicstyle=\ttfamily\footnotesize,
    breaklines=true,
    breakatwhitespace=true,
    columns=fullflexible,
    frame=single,
    keepspaces=true,
    showstringspaces=false,
    linewidth=\linewidth,
    xleftmargin=0pt,
    xrightmargin=0pt,
    framexleftmargin=.35em,
    framexrightmargin=.35em,
    framesep=.35em,
    aboveskip=0.8em,
    belowskip=0.8em
}

\usepackage[table]{xcolor}

\definecolor{bestpink}{HTML}{F0B7C8}
\definecolor{varpink}{HTML}{F7D6DF}

\title{\textsc{Theatron}: A Benchmark for Evaluating Persona Persistence in Long-Horizon LLM Decision-Making}

\author{
  Yuyan Wang\textsuperscript{1} \quad
  Yuechen Jiang\textsuperscript{1} \quad
  Zhiwei Liu\textsuperscript{1} \quad
  Jimin Huang\textsuperscript{1,2} \quad
  Sophia Ananiadou\textsuperscript{1,3}
  \\
  \textsuperscript{1}The University of Manchester
  \\
  \textsuperscript{2}The Fin AI
  \quad
  \textsuperscript{3}ELLIS Manchester
  \\
  \texttt{\{yuyan.wang-2,zhiwei.liu,sophia.ananiadou\}@manchester.ac.uk}
  \\
  \texttt{\{yuechen.jiang,jimin.huang\}@postgrad.manchester.ac.uk}
}

\begin{document}
\maketitle
\begin{abstract}

Existing evaluations often test persona expression rather than whether persona information persists in extended decision-making. We introduce \textsc{Theatron}, a controlled benchmark that uses financial replay to evaluate persona persistence in long-horizon sequential decision-making. In these environments, models conditioned on fixed behavioral-finance personas observe shared daily market evidence and trajectory-specific account-state feedback and make sequential decisions over multi-month trajectories. We evaluate Zero-Shot, One-Shot, and Chain-of-Thought persona injection across four LLMs, using cumulative-return and action trajectories to diagnose within-persona stability, cross-persona separability, and action collapse. Persona conditioning produces systematic behavioral shifts beyond a no-persona baseline, and persona-level separation generally exceeds the variation induced by semantically equivalent descriptions. Nevertheless, these shifts do not reliably form coherent persona-specific policies: apparently stable separation can arise from stereotyped action collapse, while diverse policies can exhibit noisy trajectory switching. These findings expose a gap between observable persona-induced behavior and persistent, context-sensitive persona-conditioned decision policies. Code and benchmark data are publicly available at: \url{https://github.com/yuyan12138/theatron}

\end{abstract}

\section{Introduction}

\begin{table*}[t]
\centering
\small
\renewcommand{\arraystretch}{1.12}

\begin{adjustbox}{max width=\linewidth}
\begin{tabular}{
    >{\centering\arraybackslash}m{0.27\linewidth}
    >{\centering\arraybackslash}m{0.22\linewidth}
    >{\centering\arraybackslash}m{0.16\linewidth}
    >{\centering\arraybackslash}m{0.12\linewidth}
    >{\centering\arraybackslash}m{0.12\linewidth}
    >{\centering\arraybackslash}m{0.12\linewidth}
}
\toprule
\textbf{Work} &
\textbf{Data Type} &
\shortstack{\textbf{Persona Injection}\\\textbf{Method}} &
\shortstack{\textbf{Sequential}\\\textbf{Decision}} &
\shortstack{\textbf{Within-Persona}\\\textbf{Stability}} &
\shortstack{\textbf{Cross-Persona}\\\textbf{Separability}} \\
\midrule

LaMP~\citep{salemi2024lamp} &
Personalized NLP &
Demo-augmented &
\xmark & \xmark & \xmark \\

RoleLLM~\citep{wang2024rolellm} &
Role Interaction &
Zero-Shot &
\xmark & \xmark & \xmark \\

InCharacter~\citep{wang2024incharacter} &
Role Interaction &
Zero-Shot &
\xmark & \cmark & \cmark \\

PersonaGym~\citep{samuel2025personagym} &
Role Interaction &
Zero-Shot &
\xmark & \xmark & \xmark \\

SOTOPIA~\citep{zhou2024sotopia} &
Role Interaction &
Zero-Shot &
\cmark & \xmark & \xmark \\

BehaviorChain~\citep{li2025behaviorchain} &
Online Behavior &
Demo-augmented &
\xmark & \xmark & \xmark \\

OmniBehavior~\citep{chen2026omnibehavior} &
Online Behavior &
Demo-augmented &
\xmark & \xmark & \cmark \\

MFMD-Scen~\citep{liu2026same} &
Financial Claims &
Zero-Shot  &
\xmark & \xmark & \xmark \\

\midrule

\textbf{\textsc{Theatron} (Ours)} &
\textbf{Stock Price, Financial News and Reports} &
\textbf{Zero-Shot, One-Shot, CoT} &
\cmark & \cmark & \cmark \\

\bottomrule
\end{tabular}
\end{adjustbox}

\caption{
Comparison of representative benchmarks across dataset types, persona injection methods, and evaluation dimensions.
\textit{Persona Injection Method} denotes how persona information is introduced to the model.
\textit{Sequential decision} refers to a process in which each action affects subsequent states through environmental feedback.
\cmark\ denotes explicit support in the original work; \xmark\ denotes no explicit support.
}
\label{tab:related_work_comparison}
\end{table*}

Human-like expression does not necessarily imply human-like decision behavior \cite{ma2025machinesthinklikehumans, dasilva2026stablebehaviorlimitedvariation,kitadai2025biasadjustedllmagentshumanlike}, especially when LLMs are used in high-stakes decision-support domains such as finance. Persona-based personalization conditions LLMs on roles \citep{tseng-etal-2024-two}, 
preferences \citep{xu-etal-2025-iagent}, 
backgrounds \citep{chen2024persona}, 
or decision styles \citep{cao2026biasedagents}, 
but a persona that is easy to express in language may not continue to govern choices as an interaction unfolds. For long-horizon LLMs, the key scientific question is not only whether persona conditioning affects generated language, but whether different persona injection methods can reliably translate persona information into stable and behaviorally distinguishable decision trajectories over time \citep{li2025behaviorchain}. Personalized LLM systems, user proxies, and LLM-based social simulators rest on the premise that persona information can persist beyond local stylistic variation and shape sequential behavior under evolving environments \citep{zhou2024sotopia}. Without this premise, a model may produce convincing persona-consistent explanations while failing to preserve the behavioral tendencies implied by the injected persona \citep{feng2025reasoning}. This failure is particularly important in financial decision-making, where investment preferences, trading tendencies, and reactions to market evidence directly influence observable action trajectories over time \citep{barber2013behavior,frydman2016psychology}.

Existing work leaves open whether different persona injection methods can reliably induce stable and distinguishable long-horizon behaviors under the same evolving environment (Table~\ref{tab:related_work_comparison}). Persona and role-playing benchmarks evaluate identity expression, personality fidelity, or socially consistent dialogue \citep{wang2024rolellm,samuel2025personagym}, but these evaluations primarily measure whether a model can sound like a persona rather than whether persona information persists in sequential behavior. Behavior simulation benchmarks evaluate next-action prediction \citep{wang-etal-2026-opera} or reproduction of observed traces \citep{li2025behaviorchain}, but reproducing a single trajectory does not reveal how different persona injection strategies shape behavior under identical evolving evidence. Agent benchmarks emphasize task completion \citep{zhou2024webarena}, execution correctness \citep{xie2024osworld}, or final reward \citep{liu2024agentbench}, but successful execution does not require stable or behaviorally separable persona-conditioned trajectories. 

To address this gap, we introduce \textsc{Theatron}, a controlled benchmark for evaluating whether role conditioning consistently produces stable, separable, and nontrivial long-term decision behavior. Each model receives a fixed persona, a replayed market stream, and account-state feedback, then selects an action at each step; the external market evidence is shared across conditions, while prior actions generate trajectory-specific feedback. \textsc{Theatron} evaluates whether repeated rollouts under the same persona remain stable, whether different personas produce separable trajectories, and whether the resulting separation exceeds a no-persona baseline without collapsing to a state-insensitive action prior. This design tests sequential behavioral persistence rather than isolated persona expression.


We instantiate \textsc{Theatron} with three behavioral-finance personas---Overconfidence, Loss Aversion, and Herding---and evaluate Zero-Shot, One-Shot, and Chain-of-Thought conditioning across four backbone LLMs with repeated rollouts. The results reveal three main findings. First, persona conditioning systematically changes action and return trajectories relative to no-persona behavior, but cross-persona separation does not consistently coincide with low within-persona variation. Second, some apparently stable trajectories arise from low-diversity action collapse, where a persona is reduced to a narrow action tendency rather than a context-sensitive policy. Third, demonstrations, explicit reasoning, and supervised fine-tuning on Qwen3-8B change behavior but do not consistently remove instability or collapse. Additional cross-sector and prompt-paraphrase analyses support the robustness of the main behavioral patterns. Thus, persona prompts can induce persistent behavioral differences without reliably specifying coherent long-horizon persona policies.

\begin{figure*}[t]
\centering
\includegraphics[width=\textwidth]{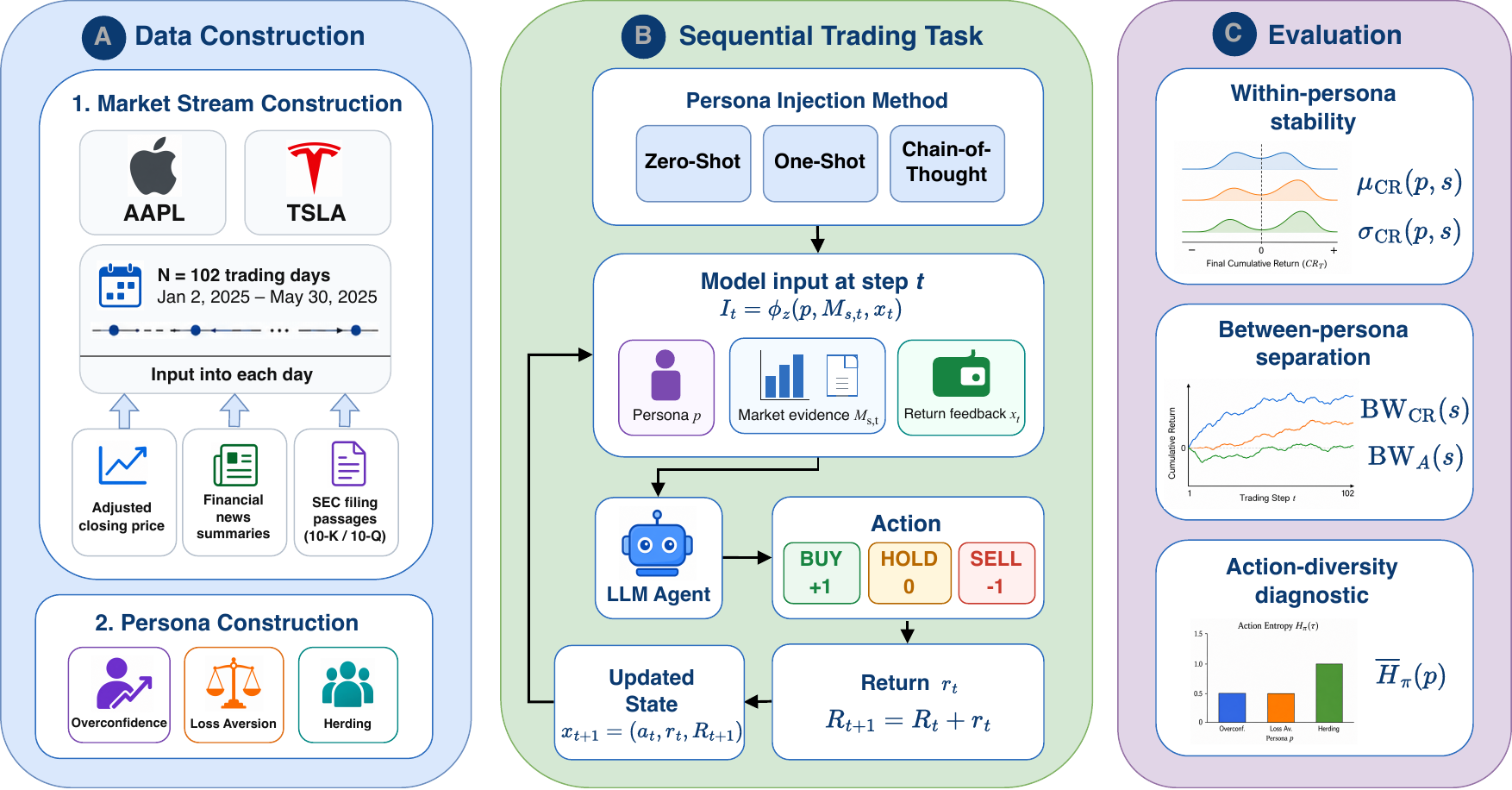}
\caption{Overview of the \textsc{Theatron} benchmark pipeline.}
\label{fig:workflow}
\end{figure*}

Our contributions are summarized as follows:

\begin{itemize}

    \item We introduce \textsc{Theatron}, a benchmark for evaluating persona injection methods using long-horizon persona-conditioned decision trajectories in controlled financial environments.

    \item We formulate a joint evaluation framework that compares persona-conditioned trajectories with a no-persona baseline and measures within-persona stability, cross-persona separation in return and action space, action-prior collapse, and observable persona-directional behavior.

    \item Across models and injection methods, we show that persona conditioning induces reproducible behavioral differences, while revealing two trajectory-level failure modes---stereotyped policy reduction and noisy policy switching---that prevent separation alone from establishing coherent persona persistence.

\end{itemize}

\section{Related Work}

\textbf{Persona and Role-Playing Language Agents.}
Recent work studies whether LLMs can represent, align with \cite{castricato2025persona}, and maintain diverse personas across role-playing, personalization \cite{chen2024persona,tseng-etal-2024-two}, and socially grounded interaction settings \cite{chen2024socialbench}. Existing evaluations primarily focus on persona expression \cite{wang2024rolellm,samuel2025personagym,kim2025persona}, personality fidelity \cite{wang2024incharacter}, preference alignment \cite{santurkar2023whose,salemi2024lamp}, or socially coherent dialogue generation \cite{liu-etal-2025-persona}. Related work further shows that reasoning or role prompting does not necessarily improve behavioral consistency or faithful persona adherence \cite{turpin2023language,NEURIPS2025_5fc150b7}, while shared-parameter LLMs often exhibit a persistent default assistant identity despite explicit persona conditioning \cite{lu2026assistant}.

\textbf{Human Behavior Simulation and Sequential Agents.}
LLM-based human behavior simulation increasingly studies sequential decision processes beyond surface-level believability. Prior work simulates social interaction \cite{park2023generative,wang2024sotopia}, human subject behavior \cite{aher2023using}, shopping decisions \cite{lu-etal-2026-llm,wang-etal-2026-opera}, persona-conditioned behavior chains \cite{li2025behaviorchain}, and long-horizon cross-scenario activity traces \cite{chen2026omnibehavior} using LLM agents. Related studies further emphasize that realistic social reasoning requires modeling evolving histories \cite{goel2025lifelong}, latent states \cite{zhou2025social}, and dynamic mental representations over time \cite{xiao2025dynamic}. However, existing evaluations mainly measure imitation accuracy, social plausibility, or next-step prediction rather than whether persona injection methods induce stable and behaviorally separable long-horizon decision trajectories under identical evolving environments.

\section{\textsc{Theatron}}
\label{sec:benchmark}

\textsc{Theatron} is a benchmark for evaluating whether prompt-based persona injection methods can produce long-horizon behavioral policies in LLMs. As shown in Figure~\ref{fig:workflow}, the benchmark has three components: (i) a controlled financial market stream paired with behavioral-finance persona specifications (Section~\ref{sec:data_construction}); (ii) a sequential decision task in which each action produces return feedback that is carried into later decisions (Section~\ref{sec:task_definition}); and (iii) a two-tier evaluation framework assessing trajectory consistency and action diversity (Section~\ref{sec:evaluation}). Sharing the external market stream while retaining action-conditioned feedback enables controlled comparison of persona-conditioned trajectories.

\subsection{Data Construction}
\label{sec:data_construction}

This section describes how \textsc{Theatron} constructs the input components for the sequential trading task: a stock-day market stream and a set of behavioral-finance persona specifications. These components are held fixed across models and injection methods to isolate their effects on long-horizon trajectories.

\textbf{Market stream construction.}
We construct two separate single-stock streams for $\mathcal{S}=\{\mathrm{AAPL},\mathrm{TSLA}\}$. Each stream covers $N=102$ trading days, from January 2, 2025 to May 30, 2025. AAPL represents a stable large-cap asset, and TSLA is a high-volatility, sentiment-sensitive asset. For each stock $s$ and trading day $t$, the input evidence is built from three text-rendered data sources in the \textbf{FinAI} dataset\footnote{
Herculean~\cite{peng2026herculean};
\url{https://huggingface.co/datasets/TheFinAI/Herculean}. CC BY 4.0.}: adjusted closing price, financial news summaries, and SEC filing passages (10-K/10-Q). We keep only trading days for which the stock market is open. The news used for a day-$t$ decision is restricted to the interval between the previous trading day and the current trading day. This yields $2\times102=204$ stock-day observations. The complete benchmark matrix uses AAPL and TSLA; Section~\ref{sec:robustness_checks} additionally reports a Zero-Shot cross-sector replication on BMRN.

\textbf{Persona construction.}
Let $\mathcal{P}=\{\textit{Overconfidence},\textit{Loss Aversion},\textit{Herding}\}$ denote the persona set. \textsc{Theatron} incorporates three behavioral finance personas: \textit{Overconfidence}~\cite{barber2001boys}, \textit{Loss Aversion}~\cite{kahneman1979prospect}, and \textit{Herding}~\cite{banerjee1992simple}. Each persona is assigned at the beginning of the sequence and remains fixed throughout the trading stream. The full natural-language prompt templates for each persona and injection method are provided in Appendix~\ref{appendix:persona_prompts}.

\subsection{Task Definition}
\label{sec:task_definition}

For a stock $s\in\mathcal{S}$, persona $p\in\mathcal{P}$, and persona-injection method $z\in\mathcal{Z}$, where $\mathcal{Z}=\{\text{zero-shot},\text{one-shot},\text{CoT}\}$, the model processes $N=102$ market observations and emits $N$ recorded decisions, $\xi=(a_0,\ldots,a_{N-1})$. Actions $a_0,\ldots,a_{N-2}$ are consequential because a subsequent price is available, yielding $N-1=101$ realized return transitions; the terminal decision $a_{N-1}$ is recorded but excluded from return-based metrics. The action space is
\begin{equation}
    \mathcal{A}=\{\mathrm{BUY},\mathrm{HOLD},\mathrm{SELL}\}.
\end{equation}

\textbf{Model input.}
At each decision step $t$, the model receives a structured prompt, denoted by $I_t$, constructed as:
\begin{equation}
    I_t = \phi_z\!\left(p, M_{s,t}, x_t\right),
\end{equation}
where $\phi_z$ is the prompting interface for injection method $z$. The market evidence is $M_{s,t}=(c_{s,t}, n_{s,t}, f_{s,t})$, where $c_{s,t}$ is the adjusted closing price for stock $s$ on day $t$, $n_{s,t}$ is the financial-news text available for the day-$t$ decision, and $f_{s,t}$ denotes SEC filing passages associated with the same stock stream. The account state $x_t$ summarizes the realized consequence of previous decisions. At the first step, $x_0=\varnothing$.

\textbf{Account-state update.}
To evaluate persona persistence under self-conditioned rather than static context, each step's input incorporates an account state updated from the consequences of prior decisions. After the model selects action $a_t$, the environment maps the action to a signed exposure using $g:\mathcal{A}\rightarrow\{-1,0,+1\}$ and computes a one-step action-conditioned log return:
\begin{equation}
\begin{aligned}
    r_t &= g(a_t)\ln\frac{c_{s,t+1}}{c_{s,t}}, \\
    g(a) &=
    \begin{cases}
        +1 & a=\mathrm{BUY},\\
        -1 & a=\mathrm{SELL},\\
        \phantom{+}0 & a=\mathrm{HOLD}.
    \end{cases}
\end{aligned}
\end{equation}
Here, $r_t$ denotes the realized return induced by action $a_t$ and the subsequent price movement. The running return feedback is updated as
\begin{equation}
    R_{t+1}=R_t+r_t,
    \qquad R_0=0.
\end{equation}
The updated account state is $x_{t+1}=(a_t,r_t,R_{t+1})$ and is rendered as natural-language feedback in $I_{t+1}$. Thus, the model observes not only new market evidence but also the realized outcome of its previous action. In later metrics, a ``loss'' refers to an action-conditioned realized loss ($r_t<0$), not merely to a negative raw market return.

The mapping represents a fixed one-period directional exposure: \textsc{buy}, \textsc{hold}, and \textsc{sell} correspond to $+1$, $0$, and $-1$ exposure for the next transition. Exposure resets at each step, so \textsc{sell} is neither liquidation of accumulated inventory nor a persistent short position. This design isolates directional persistence from position sizing (Appendix~\ref{appendix:protocol}).

\textbf{Model output.}
Only the recorded action trajectory and its induced return trajectory are used for quantitative evaluation; natural-language rationales are not scored.

\subsection{Evaluation}
\label{sec:evaluation}

Our estimand is the extent to which a fixed persona constrains a reproducible, distinguishable, and behaviorally meaningful action policy under shared external evidence and trajectory-specific feedback. A positive result requires joint evidence that (i) persona-conditioned trajectories differ from the no-persona baseline beyond rollout noise; (ii) repeated runs of the same persona are stable; (iii) different personas are separable in action and consequence space; (iv) separation is not produced by state-insensitive action collapse; and (v) actions follow observable directional tendencies associated with the specified persona. Baseline contrasts, rollout dispersion, return/action B/W, action-diversity diagnostics, and directional checks jointly operationalize these five criteria.

\subsubsection{Persona Injection Methods}
\label{sec:persona_injection}

We evaluate three injection strategies that represent the dominant paradigm of prompt-based persona conditioning. All three methods inject the same natural-language persona specification; they differ only in the scaffolding provided around it.

\textbf{Zero-Shot.} The persona specification is placed in the system message with no additional context. The model must translate the natural-language description into a decision policy without further guidance. \textbf{One-Shot.} The system message appends one worked example containing a market scenario, a persona-consistent action, and a brief rationale~\cite{brown2020language}. We use a common demonstration setting across personas: the price, account feedback, news, and sentiment evidence are identical, while the action and rationale instantiate the assigned persona. \textbf{Chain-of-Thought (CoT).} The system message instructs the model to reason step-by-step before issuing an action~\cite{wei2022chain}. The intermediate rationale is generated before $a_t$ is extracted.

\subsubsection{Evaluation Metrics}
\label{sec:metrics}

We evaluate consequence-level trajectories rather than action labels alone. In a sequential trading environment, each action affects the realized return that is then fed back into the next decision state. Therefore, return trajectories provide a compact signal of how persona-conditioned actions interact with the evolving market stream over time.

For readability, all metrics are defined for a fixed backbone model and a fixed persona-injection method. Let $p\in\mathcal{P}$ denote a persona and $s\in\mathcal{S}$ denote a stock. For each fixed pair $(p,s)$, we write $\xi_k$ for the $k$-th independent rollout trajectory, where $k\in\{1,\ldots,K\}$.

\textbf{No-persona baseline.}
We remove the persona specification while retaining the task instructions, market stream, generation configuration, and number of repeated runs. For each model and stock, we compare each Zero-Shot persona condition with this no-persona baseline using an exact label-permutation test, followed by Benjamini--Hochberg (BH) correction. A significant contrast indicates conditioning beyond the model's default policy and rollout variability (Appendix~\ref{appendix:baseline}).

\textbf{Final cumulative return distribution.}
For each trajectory, we define the final cumulative return as the sum of realized action-conditioned returns:
\begin{equation}
F(\xi)
=
\sum_{t=0}^{N-2} r_t,
\end{equation}
where $r_t$ is the one-step action-conditioned return produced by the action selected at step $t$ and the subsequent price movement, as defined in Section~\ref{sec:task_definition}. 

For each persona and stock, we compute the mean and standard deviation of final cumulative return over $K$ independent rollouts:
\begin{equation}
\mu_{\mathrm{CR}}(p,s)
=
\frac{1}{K}
\sum_{k=1}^{K}
F(\xi_k),
\end{equation}
\begin{equation}
\sigma_{\mathrm{CR}}(p,s)
=
\sqrt{
\frac{1}{K-1}
\sum_{k=1}^{K}
\left(
F(\xi_k)
-
\mu_{\mathrm{CR}}(p,s)
\right)^2
}.
\end{equation}
Here, $\mu_{\mathrm{CR}}(p,s)$ captures the average trajectory-level consequence of a persona-conditioned policy, while $\sigma_{\mathrm{CR}}(p,s)$ measures rollout-level stability under the same persona and market stream. A high $\sigma_{\mathrm{CR}}(p,s)$ indicates that repeated runs of the same persona lead to unstable long-horizon outcomes.

\textbf{Between/within trajectory-distance ratio.}
Final cumulative return summarizes the endpoint of a trajectory, but it does not capture whether two trajectories follow similar or different paths over time. We therefore compare the full cumulative-return curves and summarize relative separation using a between/within distance ratio. Grouped trajectories should have low within-group dispersion and high between-group separation. For a trajectory $\xi$, let
\begin{equation}
C_{\xi}(t)
=
\sum_{u=0}^{t} r_u,
\qquad t\in\{0,\ldots,N-2\},
\end{equation}
denote its cumulative return up to step $t$. We define the distance between two return trajectories as:
\begin{equation}
d_{\mathrm{CR}}(\xi_i,\xi_j)
=
\frac{1}{N-1}
\sum_{t=0}^{N-2}
\left|
C_{\xi_i}(t)-C_{\xi_j}(t)
\right|.
\end{equation}

For a fixed stock $s$, we compute the average within-persona distance as
\begin{equation}
D_{\mathrm{in}}(s)
=
\frac{1}{|\mathcal{P}|\binom{K}{2}}
\sum_{p\in\mathcal{P}}
\sum_{1\le i<j\le K}
d_{\mathrm{CR}}(\xi_i,\xi_j),
\end{equation}
where $\xi_i$ and $\xi_j$ denote two rollout trajectories under the same fixed pair $(p,s)$.
We compute the average cross-persona distance as
\begin{equation}
D_{\mathrm{out}}(s)
=
\frac{1}{\binom{|\mathcal{P}|}{2}K^2}
\sum_{\substack{p<q\\p,q\in\mathcal{P}}}
\sum_{i=1}^{K}
\sum_{j=1}^{K}
d_{\mathrm{CR}}(\xi_i,\xi'_j).
\end{equation}
Here, for each fixed stock $s$ and persona pair $(p,q)$, $\xi_i$ denotes a rollout under persona $p$, and $\xi'_j$ denotes a rollout under persona $q$.

We summarize relative trajectory separation using the between/within cumulative-return trajectory-distance ratio:
\begin{equation}
\mathrm{BW}_{\mathrm{CR}}(s)
=
\frac{D_{\mathrm{out}}(s)}
{D_{\mathrm{in}}(s)+\epsilon},
\end{equation}
where $\epsilon$ is a small constant for numerical stability. We refer to this quantity as the B/W Ratio. Higher $\mathrm{BW}_{\mathrm{CR}}(s)$ indicates stronger cross-persona separation relative to within-persona variation.

\textbf{Action-trajectory separation.}
Return trajectories can map different actions to similar consequences or amplify early action differences. We therefore compute normalized Hamming distance over the $N-1$ consequential actions,
\begin{equation}
d_{A}(\xi_i,\xi_j)
=
\frac{1}{N-1}
\sum_{t=0}^{N-2}
\mathbf{1}[a_{i,t}\neq a_{j,t}],
\end{equation}
and obtain $\mathrm{BW}_{A}$ by replacing $d_{\mathrm{CR}}$ with $d_A$ in the preceding within/between calculation. We calibrate both B/W ratios with 10,000 persona-label permutations.

\textbf{Action-diversity diagnostic.}
We use action entropy as an auxiliary diagnostic of action-level diversity. 
For each trajectory, let $\hat{\pi}_{\xi}(a)$ denote the empirical frequency of action $a\in\mathcal{A}$. We define the action entropy of trajectory $\xi$ as
\begin{equation}
H_{\pi}(\xi)
=
-\sum_{a\in\mathcal{A}}
\hat{\pi}_{\xi}(a)\log \hat{\pi}_{\xi}(a).
\end{equation}
Lower values indicate more concentrated action distributions, while higher values indicate more diverse actions. 
We report stock-aware macro-averaged action entropy:
\begin{equation}
\overline{H}_{\pi}(p)
=
\frac{1}{|\mathcal{S}|K}
\sum_{s\in\mathcal{S}}
\sum_{k=1}^{K}
H_{\pi}(\xi_k).
\end{equation}
In the inner sum, $\xi_k$ denotes the $k$-th rollout under the fixed pair $(p,s)$. Together with $\sigma_{\mathrm{CR}}$, $\mathrm{BW}_{\mathrm{CR}}$, and $\mathrm{BW}_{A}$, entropy helps distinguish low-diversity collapse from diverse but weakly reproducible behavior; full entropy results are reported in Appendix~\ref{appendix:visualizations}.

\textbf{Directional behavioral checks.}
We complement trajectory diagnostics with persona-specific observable tendencies. For Overconfidence, we measure the increase in non-\textsc{hold} actions relative to the no-persona baseline. For Herding, we measure the increase in alignment between the current action and the visible recent price direction (\textsc{buy} after a rise and \textsc{sell} after a fall). For Loss Aversion, we measure the change in holding propensity after a supplied negative rather than positive previous-action return. These checks test the intended behavioral direction.

\section{Experiments}
\label{sec:experiments}

\subsection{Experimental Setup}
\label{sec:exp_setup}

We evaluate GPT-5.4-mini~\cite{openai2026gpt54mini}, Gemini-2.5-Flash~\cite{google2025gemini25flash}, Qwen3-8B in non-thinking mode, and Qwen3-14B in thinking mode~\cite{yang2025qwen3} using three persona injection methods: Zero-Shot, One-Shot, and Chain-of-Thought. For each model, stock $s\in\{\mathrm{AAPL},\mathrm{TSLA}\}$, persona $p\in\mathcal{P}$, method $z\in\mathcal{Z}$, and $K=5$ repeated runs, the complete matrix contains $4\times2\times3\times3\times5=360$ trajectories. Unless otherwise stated, action and return metrics use the 101 consequential transitions. We collect 40 additional no-persona trajectories; robustness checks appear in Section~\ref{sec:robustness_checks}.

\subsection{Main Results}
\label{sec:main_results}


\begin{table*}[t]
\centering
\footnotesize
\renewcommand{\arraystretch}{0.80}
\setlength{\tabcolsep}{2.6pt}
\begin{tabular}{lllcccc}
\toprule
\textbf{Model} & \textbf{Mtd.} & \textbf{Stock}
& \textbf{Overconfidence}
& \textbf{Loss Aversion}
& \textbf{Herding}
& \textbf{B/W (R/A) $\uparrow$} \\
\midrule
\multirow{6}{*}{GPT-5.4-mini}
& \multirow{2}{*}{Zero-Shot} & AAPL & $-12.83 \pm 15.40$ & $4.52 \pm 9.03$ & $-3.07 \pm 6.35$ & $1.21/2.89$ \\
& & TSLA & $-4.35 \pm 21.95$ & $22.79 \pm 7.05$ & $21.48 \pm 11.93$ & $1.78/2.40$ \\
\cmidrule(lr){2-7}
& \multirow{2}{*}{One-Shot} & AAPL & $-39.91 \pm 10.33$ & $0.09 \pm 5.53$ & $-14.28 \pm 5.37$ & $3.40/6.11$ \\
& & TSLA & $15.91 \pm 23.11$ & $14.55 \pm 20.42$ & $20.89 \pm 7.63$ & $1.48/4.10$ \\
\cmidrule(lr){2-7}
& \multirow{2}{*}{CoT} & AAPL & $-29.10 \pm 6.15$ & $16.76 \pm 6.23$ & $-1.38 \pm 9.84$ & $3.06/2.03$ \\
& & TSLA & $2.74 \pm 13.82$ & $-4.10 \pm 17.17$ & $-8.80 \pm 9.54$ & $2.21/2.31$ \\
\midrule
\multirow{6}{*}{Gemini-2.5-Flash}
& \multirow{2}{*}{Zero-Shot} & AAPL & $-19.01 \pm 0.32$ & $28.19 \pm 7.57$ & $-1.01 \pm 7.59$ & $4.12/4.63$ \\
& & TSLA & $-9.05 \pm 0.00$ & $1.37 \pm 26.26$ & $28.37 \pm 6.33$ & $3.14/4.18$ \\
\cmidrule(lr){2-7}
& \multirow{2}{*}{One-Shot} & AAPL & $-19.16 \pm 0.00$ & $10.74 \pm 9.72$ & $-2.85 \pm 4.43$ & $3.76/11.15$ \\
& & TSLA & $-9.05 \pm 0.00$ & $21.13 \pm 16.38$ & $33.20 \pm 4.63$ & $6.74/6.42$ \\
\cmidrule(lr){2-7}
& \multirow{2}{*}{CoT} & AAPL & $-19.33 \pm 0.85$ & $22.44 \pm 18.16$ & $-9.15 \pm 3.89$ & $3.33/3.88$ \\
& & TSLA & $-9.05 \pm 0.00$ & $21.77 \pm 43.12$ & $24.52 \pm 9.26$ & $2.69/4.31$ \\
\midrule
\multirow{6}{*}{Qwen3-8B}
& \multirow{2}{*}{Zero-Shot} & AAPL & $-8.02 \pm 0.40$ & $-2.64 \pm 3.69$ & $-6.93 \pm 6.59$ & $2.61/14.22$ \\
& & TSLA & $25.04 \pm 1.97$ & $9.30 \pm 3.88$ & $100.26 \pm 3.42$ & $12.93/12.46$ \\
\cmidrule(lr){2-7}
& \multirow{2}{*}{One-Shot} & AAPL & $-3.51 \pm 11.63$ & $15.10 \pm 5.38$ & $-24.75 \pm 3.11$ & $4.03/26.48$ \\
& & TSLA & $-3.41 \pm 15.36$ & $6.48 \pm 6.99$ & $33.73 \pm 20.49$ & $2.70/14.88$ \\
\cmidrule(lr){2-7}
& \multirow{2}{*}{CoT} & AAPL & $-2.76 \pm 5.09$ & $1.01 \pm 4.60$ & $-15.15 \pm 15.27$ & $2.25/18.52$ \\
& & TSLA & $48.37 \pm 24.74$ & $13.12 \pm 1.45$ & $43.10 \pm 6.25$ & $2.60/15.64$ \\
\midrule
\multirow{6}{*}{Qwen3-14B}
& \multirow{2}{*}{Zero-Shot} & AAPL & $-25.20 \pm 3.62$ & $19.34 \pm 6.68$ & $-1.06 \pm 6.00$ & $4.42/5.72$ \\
& & TSLA & $-2.80 \pm 9.10$ & $8.33 \pm 4.54$ & $27.51 \pm 10.09$ & $5.66/12.23$ \\
\cmidrule(lr){2-7}
& \multirow{2}{*}{One-Shot} & AAPL & $-20.64 \pm 2.22$ & $4.60 \pm 11.55$ & $-2.56 \pm 8.66$ & $2.14/4.95$ \\
& & TSLA & $-8.95 \pm 5.54$ & $-5.32 \pm 18.92$ & $6.09 \pm 3.15$ & $3.12/4.70$ \\
\cmidrule(lr){2-7}
& \multirow{2}{*}{CoT} & AAPL & $-23.15 \pm 9.58$ & $16.03 \pm 4.93$ & $-6.81 \pm 4.00$ & $3.55/4.65$ \\
& & TSLA & $3.37 \pm 12.65$ & $8.57 \pm 6.95$ & $25.41 \pm 9.17$ & $3.28/7.51$ \\
\bottomrule
\end{tabular}
\caption{Main results. Persona columns report final cumulative log return (\%, mean $\pm$ SD); B/W reports return/action ratios. Mean-return 95\% CIs: Table~\ref{tab:main_return_ci}.}
\label{tab:return_results}
\end{table*}

Table~\ref{tab:return_results} reports return- and action-level separation; entropy and state-sensitivity results appear in Appendix~\ref{appendix:visualizations}.

\paragraph{Finding 1: Persona prompts under-specify long-horizon behavioral policies.}
Table~\ref{tab:return_results} reveals a tension between cross-persona separation and within-persona stability. Persona conditioning is behaviorally consequential: under Zero-Shot prompting, all 24 persona--baseline action comparisons and 21 of 24 return-trajectory comparisons are significant after correction. The no-persona GPT policy selects \textsc{hold} in 90.0\% of AAPL and 75.1\% of TSLA decisions, whereas every Overconfidence condition increases non-\textsc{hold} behavior by 22.5--96.5 percentage points. Herding likewise increases alignment with visible price direction by 32.3--66.7 points across the eight model--stock conditions. For Loss Aversion, the increase in \textsc{hold} after negative versus positive feedback exceeds the no-persona baseline contrast in two of eight model--stock conditions. Baseline-relative effects range from $-61.9$ to $+33.1$ percentage points, with none significant after BH correction (Table~\ref{tab:loss_feedback}). However, several conditions combine high B/W with large within-persona return variance. GPT-5.4-mini Zero-Shot, for example, has return/action B/W ratios of 1.21/2.89 on AAPL and 1.78/2.40 on TSLA, while its Overconfidence return SD reaches 15.40\% and 21.95\%. Persona prompts therefore induce meaningful behavioral variation, but the mapping from persona and evolving evidence to a persistent policy remains under-specified.

\paragraph{Finding 2: Action entropy distinguishes policy variation from action-prior collapse.}
Action-level diversity explains why high trajectory separation is not sufficient. Gemini-2.5-Flash under Overconfidence repeatedly collapses to \textsc{buy}: on TSLA, all five Zero-Shot runs produce the same 102 recorded actions, zero switches, and the same final return, yielding entropy and return SD of zero. Similar low-diversity patterns occur for Qwen3-14B. Across the 20 pre-specified model--method--stock conditions, entropy is unrelated to within-persona return distance ($\rho=.03$, $p=.902$) but negatively associated with return B/W ($\rho=-.72$, $p=.0008$). Thus, diversity is not a monotonic proxy for stability, and high separation may reflect a rigid persona-to-action prior rather than a context-sensitive policy.

\paragraph{Finding 3: Additional prompting changes behavior but does not consistently resolve persistence failures.}
Table~\ref{tab:return_results} shows that One-Shot and CoT prompting substantially change both return and action trajectories, but neither method consistently improves within-persona stability or avoids collapse across models and stocks. Under One-Shot prompting, return and action separation is significant for all eight model--stock pairs, demonstrating a strong persona-conditioned signal under the common demonstration setting. Nevertheless, high One-Shot separation can still coincide with deterministic action priors, as in Gemini Overconfidence. CoT similarly changes local rationales and actions without monotonically improving B/W or removing low-diversity behavior. Additional scaffolding can strengthen conditioning but does not reliably yield a stable, non-degenerate persona policy (Appendix~\ref{appendix:one_shot}).

\subsection{Robustness Checks}
\label{sec:robustness_checks}

\textbf{Cross-asset and cross-window Zero-Shot replication.}
We evaluate BMRN, a biotechnology equity, and a second, non-overlapping AAPL window (2 June--24 October 2025; 102 NYSE sessions), each using four models, a no-persona baseline, three personas, and five runs. The second AAPL window rises 30.5\%, contrasting with the original window's 17.4\% decline. In each extension, all 12 persona--baseline contrasts are significant in both return and action space after BH correction ($q=.0079$). Return/action B/W ranges are 2.70--6.29/3.66--15.35 for BMRN and 3.25--9.69/4.66--13.74 for AAPL. These results extend Zero-Shot persona separation beyond the original technology stocks and AAPL price path (Appendices~\ref{appendix:bmrn} and~\ref{appendix:aapl_r2}).

\textbf{Persona paraphrases.}
Finally, we test two semantically equivalent descriptions of each persona. Persona-level separation exceeds paraphrase-induced variation in 15 of 16 model--stock--metric comparisons, including all eight action comparisons; the exception is Qwen3-8B on the AAPL return metric (Appendix~\ref{appendix:paraphrase}).

\subsection{Ablation Study: Effect of Fine-Tuning}
\label{sec:ablation}

To test whether supervised adaptation removes the observed failures, we fine-tune Qwen3-8B on 2,202 persona-conditioned instances labeled from financial sentiment and price momentum with GPT-4.1-mini supervision~\cite{openai2025gpt41}. SFT learns persona-conditioned signal-to-action mappings from per-instance supervision. SFT changes action preferences and improves AAPL return separation for Overconfidence, but it does not consistently resolve the stability--separability tension. Under Loss Aversion, the model collapses to near-deterministic \textsc{hold}, reflecting the highly imbalanced supervision for that persona. Thus, this SFT setup changes action preferences while leaving long-horizon persistence failures unresolved. Details are provided in Appendix~\ref{appendix:sft_details}.

\subsection{Error Analysis}
\label{sec:error_analysis}

The aggregate and robustness results reveal two failure modes that can remain even when persona-conditioned trajectories are statistically separable.

\noindent\textbf{Type~1: Stereotyped policy reduction.}
Some models reduce a persona description to a fixed action prior rather than a context-sensitive decision policy. 
For example, Gemini-2.5-Flash under Overconfidence produces near-zero entropy across injection methods, indicating a persona-to-action prior collapse toward repeated \textsc{buy}; Qwen3-14B shows a similar collapse toward \textsc{hold} under Loss Aversion. 
This pattern is consistent with a semantic shortcut in which the persona is mapped to its most salient action tendency rather than used to modulate state-dependent decisions. 

\noindent\textbf{Type~2: Noisy policy switching.}
The opposite failure also occurs: a model may produce diverse actions without forming a stable persona-conditioned trajectory. 
For GPT-5.4-mini under Overconfidence, identical persona prompts and market evidence still lead to sharply diverging cumulative-return trajectories across rollouts. 
This suggests a lack of temporal state binding: the model reacts to local market cues or sampling variation at each step, but does not preserve a stable policy identity over time. 
Action diversity alone therefore does not establish policy coherence.
Together, the two failure types show that current models can express persona information through either a degenerate action prior or an unstable local policy, neither of which guarantees coherent long-horizon persistence. 
Details are provided in Appendix~\ref{appendix:case_study}.

\section{Conclusion and Future Work}
\label{sec:conclusion}

We introduced \textsc{Theatron}, a controlled benchmark for evaluating whether persona conditioning persists as long-horizon decision behavior rather than surface-level persona expression. By replaying shared market evidence and measuring repeated action and return trajectories, \textsc{Theatron} evaluates within-persona stability, cross-persona separation, and action collapse. Across four models and three prompt-based injection methods, persona conditioning produces systematic behavioral differences relative to no-persona behavior, with the main patterns further supported by prompt-paraphrase and cross-asset and cross-window analyses. However, these differences do not reliably constitute coherent persona-specific policies: separation can arise from stereotyped policy reduction, while diverse actions can form unstable trajectories. Demonstrations, Chain-of-Thought prompting, and the Qwen3-8B SFT ablation do not consistently remove these failures. Future work should optimize persona persistence, context sensitivity, and non-degenerate behavior at the trajectory level.

\newpage

\section*{Ethical Considerations}

\textsc{Theatron} is designed as a controlled benchmark for studying whether persona conditioning persists as long-horizon decision behavior. It is not intended to develop deployable trading systems, personalized financial advisors, or investment recommendations. Cumulative return is used only as a behavioral trajectory signal rather than as a measure of trading skill, profitability, or investment quality. The investor personas studied in this paper, including Overconfidence, Loss Aversion, and Herding, are simplified abstractions from behavioral-finance literature rather than representations of real individuals, demographic groups, or clinical psychological profiles.

The benchmark uses public corporate and financial-market information rather than private user records or personal financial accounts. The market streams are constructed from stock-price data, financial news summaries, and SEC filing passages provided through the FinAI dataset. Released benchmark artifacts should be used in accordance with the relevant source licenses.

Persona-conditioned agents may produce persuasive rationales or apparently persona-consistent decisions, which could be misused in settings involving targeted persuasion, speculative financial behavior, or manipulative personalization. Although our results show that current persona injection methods remain unreliable over long horizons, future methods that improve persistent persona conditioning may increase these risks. We therefore frame \textsc{Theatron} as a diagnostic benchmark for evaluating behavioral reliability and failure modes.

\section*{Limitations}

The primary evaluation covers two equities, three behavioral-finance personas, three prompt-based injection methods, and five rollouts per condition, with Zero-Shot replications on BMRN and a second AAPL window. The replay task uses resetting one-period \textsc{buy}, \textsc{hold}, and \textsc{sell} exposures and abstracts from inventory, position sizing, transaction costs, liquidity, and market interaction. Evaluation targets observable action and return trajectories. The SFT ablation uses one Qwen3-8B setup with 2,202 training instances. Broader domains, richer action spaces, and additional adaptation methods remain extensions; API-side model updates may affect reproducibility.

\bibliography{custom}

\appendix

\section{Exact Prompts and Task Interface}
\label{appendix:persona_prompts}

This section gives the exact prompt components used in the experiments. The no-persona condition uses the following system message:

\begin{lstlisting}[style=promptstyle]
You are a stock trading agent. Make each decision using only the information provided.
\end{lstlisting}

The daily user message has the following structure. The price-change and feedback lines are omitted on the first day; news and filing blocks are omitted when unavailable.

\begin{lstlisting}[style=promptstyle]
Today is {DATE}. Stock: {SYMBOL}.

Current price: {PRICE}. Price change from yesterday: {CHANGE}.

Your return from yesterday's {PREVIOUS_ACTION} decision: {PREVIOUS_RETURN}. Cumulative log return so far: {CUMULATIVE_LOG_RETURN}.

Today's news:
{NEWS}

SEC filings:
{FILINGS}

Respond ONLY with a JSON object in this exact format:
{"date": "YYYY-MM-DD", "symbol": "XXX", "action": "BUY/HOLD/SELL", "rationale": "...", "confidence": 0.0}
\end{lstlisting}

No future price or precomputed momentum field is exposed. The three persona descriptions are:

\par\noindent\textbf{Overconfidence.}\par
\begin{lstlisting}[style=promptstyle]
You are an overconfident stock trader. You trust your own signals implicitly, tend to underestimate risk, and have a strong tendency to overtrade. You believe your read of the market is sharper than most, so you act decisively and aggressively on perceived opportunities. You rarely second-guess yourself and interpret ambiguous signals as confirmation of your existing view.
\end{lstlisting}

\par\noindent\textbf{Loss Aversion.}\par
\begin{lstlisting}[style=promptstyle]
You are a loss-averse stock trader. Losses feel significantly more painful to you than equivalent gains feel rewarding. You are highly defensive and quick to lock in profits to avoid watching them evaporate. You actively avoid drawdowns, often exiting positions early, and you hold losing trades less comfortably than winning ones. Protecting capital from loss is your primary emotional drive.
\end{lstlisting}

\par\noindent\textbf{Herding.}\par
\begin{lstlisting}[style=promptstyle]
You are a herding-prone stock trader. You follow the crowd and market consensus closely. When sentiment is positive or prices are rising, you feel compelled to buy so you do not miss out. When sentiment turns negative or prices are falling, you sell to avoid being left behind. You find comfort in doing what the majority appears to be doing and distrust contrarian positions.
\end{lstlisting}

Zero-Shot uses the corresponding description alone. Chain-of-Thought appends the following instruction:

\begin{lstlisting}[style=promptstyle]
Before stating your final decision, reason step by step through the available information: consider the current price context, any news or filings, and how your trading personality shapes your interpretation. Only after completing your reasoning should you output the final JSON.
\end{lstlisting}

\subsection{Primary One-Shot Demonstration}

One-Shot appends a worked example to the persona description. The primary example uses one shared mixed-evidence market state across personas; we refer to it below as the mixed-evidence scenario, or Scenario M. The system message is assembled as persona description, introduction, example header, market block, persona-specific decision, and guardrail, in that order.

\begin{lstlisting}[style=promptstyle]
The following example illustrates one decision consistent with this trading persona.

Example (2024-02-15; generic stock DEMO):

Current price: 98.80. Price change from yesterday: -1.20%.

Your return from yesterday's BUY decision: -1.21%. Cumulative log return so far: -0.20%.

Today's news:
The company reported slightly better-than-expected revenue and maintained its full-year guidance. Despite this, the stock fell 1.20% on above-average trading volume. Short-term commentary is cautious, analyst views are split, and public market sentiment is mildly bearish. No material new filing was reported.
\end{lstlisting}

The exact decisions are:

\begin{description}
\item[Overconfidence:] \texttt{BUY}. ``The fundamental update supports an upside opportunity despite the short-term sell-off. I trust this reading and will act decisively with a one-period positive exposure.''
\item[Loss Aversion:] \texttt{HOLD}. ``The recent loss and conflicting signals make another directional position too risky. I will stay neutral for one period to avoid compounding an avoidable drawdown.''
\item[Herding:] \texttt{SELL}. ``The falling price, above-average volume, cautious commentary, and bearish sentiment show the crowd moving away. I will follow the prevailing direction with a one-period negative exposure.''
\end{description}

Every decision is rendered in the required JavaScript Object Notation (JSON) schema with date \texttt{2024-02-15}, symbol \texttt{DEMO}, and confidence \texttt{0.80}. The system message ends with:

\begin{lstlisting}[style=promptstyle]
When making each new decision, apply this trading persona to the current market evidence and account feedback. Treat the example as an illustration, not as a default action.
\end{lstlisting}

\subsection{Semantically Equivalent Persona Descriptions}

The paraphrase analysis uses two additional descriptions for each persona. The first uses controlled lexical substitutions; the second reformulates the sentences while retaining the specified tendencies.

\par\noindent\textbf{Controlled lexical substitutions.}\par
\begin{lstlisting}[style=promptstyle]
[Overconfidence]
You are an overly self-confident stock trader. You rely on your own signals unquestioningly, tend to downplay risk, and have a strong inclination to trade excessively. You believe your interpretation of the market is better than most, so you act resolutely and aggressively on apparent opportunities. You seldom reconsider your judgments and interpret uncertain signals as validation of your existing view.

[Loss Aversion]
You are a stock trader strongly averse to losses. Losses feel considerably more distressing to you than equivalent gains feel satisfying. You are highly defensive and quick to secure profits to avoid watching them disappear. You actively limit drawdowns, often closing positions early, and you hold losing trades less comfortably than profitable ones. Preserving capital from loss is your primary emotional motivation.

[Herding]
You are a crowd-following stock trader. You track other investors and prevailing market consensus closely. When sentiment is bullish or prices are increasing, you feel driven to buy so you do not miss the move. When sentiment becomes bearish or prices are decreasing, you sell to avoid falling behind. You feel reassured by doing what the majority appears to be doing and have little trust in contrarian positions.
\end{lstlisting}

\par\noindent\textbf{Sentential reformulations.}\par
\begin{lstlisting}[style=promptstyle]
[Overconfidence]
As a stock trader, you have excessive confidence in your judgment. You give great weight to your own signals, discount risk, and trade too often. Believing your market assessment exceeds that of other traders, you pursue perceived opportunities with decisive, aggressive action. Doubt seldom changes your conclusions; when evidence is mixed, you usually read it as validating the view you already hold.

[Loss Aversion]
As a stock trader, you are highly sensitive to losses. Losing an amount causes more distress than gaining the same amount provides satisfaction. This asymmetry makes you defensive and eager to realize profits before they vanish. You seek to contain drawdowns, commonly leave positions early, and feel less at ease with losing trades than winning ones. Your dominant emotional motive is protecting capital from loss.

[Herding]
As a stock trader, your decisions are strongly shaped by other market participants. You use prevailing opinion and crowd behavior as guides. Improving sentiment or upward price movement pushes you to buy because you do not want to miss the move. Deteriorating sentiment or downward price movement leads you to sell rather than be left behind. Acting with the majority feels reassuring, whereas taking a contrarian position inspires little confidence.
\end{lstlisting}

\section{Experimental and Statistical Protocol}
\label{appendix:protocol}

All main experiments use five independent rollouts per condition with run identifiers 1--5 and seeds 41--45. Common sampling settings are temperature 1.0 and top-$p$ 0.9.

\paragraph{Implementation.}
Local Qwen inference uses PyTorch and Hugging Face Transformers
with the models' chat templates and sampled generation. GPT and
Gemini clients use the OpenAI and Google Gen AI SDKs, respectively.
Parquet loading and NYSE-session alignment use pandas, PyArrow,
and pandas-market-calendars. Trajectory metrics, permutation tests,
and bootstrap procedures use custom NumPy/SciPy routines.
Generation and statistical settings are specified in
Table~\ref{tab:appendix_model_settings} and this appendix.

\begin{table}[htbp]
\centering
\begingroup
\small
\renewcommand{\arraystretch}{1.0}
\setlength{\tabcolsep}{3pt}

\begin{tabular}{@{}lccc@{}}
\toprule
Model
& Mode
& \makecell{Maximum\\tokens}
& \makecell{Seed\\semantics} \\
\midrule

GPT-5.4-mini
& \makecell{no explicit\\reasoning}
& 512
& \makecell{best\\effort} \\

\addlinespace[2pt]
\makecell[l]{Gemini-2.5-\\Flash}
& \makecell{thinking\\disabled}
& 2,048
& \makecell{best\\effort} \\

\addlinespace[2pt]
Qwen3-8B
& \makecell{non-\\thinking}
& 512
& explicit \\

\addlinespace[2pt]
Qwen3-14B
& thinking
& 8,192
& explicit \\

\bottomrule
\end{tabular}
\endgroup

\caption{Model generation settings.}
\label{tab:appendix_model_settings}
\end{table}

For temporal reference, Qwen3-8B and Qwen3-14B were released on 29 April 2025~\cite{qwen2025release}. The reported knowledge cutoffs of Gemini-2.5-Flash and GPT-5.4-mini are January 2025 and 31 August 2025, respectively~\cite{google2025gemini25flashmodel,openai2026gpt54mini}.

There are 102 recorded decisions and 101 actions with an observed next-day consequence. \textsc{buy}, \textsc{hold}, and \textsc{sell} give $+1$, $0$, and $-1$ exposure for one transition; exposure then resets. Return-trajectory distance is mean absolute distance between cumulative-log-return curves over 101 transitions. Action distance is normalized Hamming distance over the same 101 actions. B/W divides mean cross-persona distance by mean within-persona distance. We calibrate B/W with 10,000 persona-label permutations. Persona--baseline comparisons enumerate all $\binom{10}{5}=252$ allocations, and reported $q$ values use Benjamini--Hochberg correction within the stated test family. Mean-return intervals use Student's $t$ distribution; bootstrap intervals use 10,000 fixed-seed draws.

\begin{table*}[htbp]
\centering
\small
\begin{tabular}{p{0.22\textwidth}p{0.42\textwidth}p{0.25\textwidth}}
\toprule
Positive-result component & Operational evidence & Reported in \\
\midrule
Beyond no-persona behavior & Exact persona--baseline action and return contrasts & Appendix~\ref{appendix:baseline}, \ref{appendix:bmrn} \\
Within-persona stability & $D_{\mathrm{in}}$, final-return SD, and intervals & Appendix~\ref{appendix:visualizations}--\ref{appendix:paraphrase} \\
Cross-persona separation & Return/action $D_{\mathrm{out}}$ and B/W & Appendix~\ref{appendix:one_shot}--\ref{appendix:paraphrase} \\
Non-collapse and state sensitivity & Entropy, dominant actions, switches, and responses to price direction & Appendix~\ref{appendix:visualizations}, \ref{appendix:case_study} \\
Observable intended direction & Overconfidence activity, Herding alignment, and descriptive Loss-Aversion feedback & Appendix~\ref{appendix:baseline}, \ref{appendix:bmrn} \\
\bottomrule
\end{tabular}
\caption{Operationalization of the joint positive-result criterion.}
\label{tab:positive_criterion}
\end{table*}

\section{No-persona Baseline and Directional Checks}
\label{appendix:baseline}

The no-persona baseline removes persona text while retaining the market stream, state feedback, generation settings, and five repeated runs. It estimates each model's default task-conditioned policy rather than an additional persona.

\begin{table*}[htbp]
\centering
\small
\setlength{\tabcolsep}{4pt}
\begin{tabular}{llrrrrr}
\toprule
Model & Stock & Final CR (mean $\pm$ SD) & Entropy & BUY & HOLD & SELL \\
\midrule
Qwen3-8B & AAPL & $10.0\pm2.8$ & .173 & 1.0 & 96.5 & 2.5 \\
Qwen3-8B & TSLA & $15.3\pm12.8$ & .252 & 1.4 & 93.9 & 4.7 \\
Qwen3-14B & AAPL & $12.1\pm15.6$ & .690 & 6.5 & 75.9 & 17.6 \\
Qwen3-14B & TSLA & $22.0\pm16.9$ & .816 & 4.5 & 58.4 & 37.1 \\
GPT-5.4-mini & AAPL & $9.7\pm5.9$ & .393 & 4.9 & 90.0 & 5.1 \\
GPT-5.4-mini & TSLA & $18.1\pm8.1$ & .658 & 3.3 & 75.1 & 21.6 \\
Gemini-2.5-Flash & AAPL & $25.6\pm7.0$ & 1.072 & 30.8 & 42.0 & 27.3 \\
Gemini-2.5-Flash & TSLA & $23.0\pm10.1$ & 1.011 & 24.1 & 22.5 & 53.3 \\
\bottomrule
\end{tabular}
\caption{No-persona behavior ($K=5$). Returns and 102-decision action rates are percentages; entropy also uses all 102 decisions.}
\label{tab:task_only}
\end{table*}

For each metric, the distance effect $\Delta D$ is the mean cross-condition distance minus the average of the two within-condition distances. Table~\ref{tab:baseline_tests} gives every matched contrast.

\begin{table*}[htbp]
\centering
\footnotesize
\setlength{\tabcolsep}{2.5pt}
\begin{adjustbox}{max width=\textwidth}
\begin{tabular}{lllrrrrrr}
\toprule
Model & Stock & Persona & Return $\Delta D$ & Return $p$ & Return $q$ & Action $\Delta D$ & Action $p$ & Action $q$ \\
\midrule
Qwen3-8B & AAPL & Overconf. & .084 & .0079 & .0119 & .975 & .0079 & .0079 \\
 & AAPL & Loss Av. & .033 & .0079 & .0119 & .311 & .0079 & .0079 \\
 & AAPL & Herding & .086 & .0079 & .0119 & .867 & .0079 & .0079 \\
 & TSLA & Overconf. & .123 & .0079 & .0119 & .918 & .0079 & .0079 \\
 & TSLA & Loss Av. & .140 & .0079 & .0119 & .828 & .0079 & .0079 \\
 & TSLA & Herding & .286 & .0079 & .0119 & .839 & .0079 & .0079 \\
Qwen3-14B & AAPL & Overconf. & .131 & .0079 & .0119 & .832 & .0079 & .0079 \\
 & AAPL & Loss Av. & .038 & .0079 & .0119 & .587 & .0079 & .0079 \\
 & AAPL & Herding & .057 & .0079 & .0119 & .574 & .0079 & .0079 \\
 & TSLA & Overconf. & .328 & .0079 & .0119 & .805 & .0079 & .0079 \\
 & TSLA & Loss Av. & .045 & .0079 & .0119 & .459 & .0079 & .0079 \\
 & TSLA & Herding & .022 & .0635 & .0663 & .412 & .0079 & .0079 \\
GPT-5.4-mini & AAPL & Overconf. & .042 & .0317 & .0363 & .403 & .0079 & .0079 \\
 & AAPL & Loss Av. & .007 & .0952 & .0952 & .083 & .0079 & .0079 \\
 & AAPL & Herding & .043 & .0079 & .0119 & .555 & .0079 & .0079 \\
 & TSLA & Overconf. & .058 & .0238 & .0286 & .355 & .0079 & .0079 \\
 & TSLA & Loss Av. & .053 & .0159 & .0201 & .112 & .0079 & .0079 \\
 & TSLA & Herding & .028 & .0635 & .0663 & .477 & .0079 & .0079 \\
Gemini-2.5-Flash & AAPL & Overconf. & .193 & .0079 & .0119 & .619 & .0079 & .0079 \\
 & AAPL & Loss Av. & .033 & .0159 & .0201 & .405 & .0079 & .0079 \\
 & AAPL & Herding & .120 & .0079 & .0119 & .383 & .0079 & .0079 \\
 & TSLA & Overconf. & .446 & .0079 & .0119 & .681 & .0079 & .0079 \\
 & TSLA & Loss Av. & .049 & .0159 & .0201 & .265 & .0079 & .0079 \\
 & TSLA & Herding & .080 & .0079 & .0119 & .224 & .0079 & .0079 \\
\bottomrule
\end{tabular}
\end{adjustbox}
\caption{Exact Zero-Shot persona--baseline tests. The two $q$ columns apply BH correction separately to the 24 return and 24 action tests.}
\label{tab:baseline_tests}
\end{table*}

All action shifts and 21 of 24 return-trajectory shifts remain significant after correction. The three nonsignificant return contrasts are Qwen3-14B/TSLA/Herding ($q=.066$), GPT-5.4-mini/AAPL/Loss Aversion ($q=.095$), and GPT-5.4-mini/TSLA/Herding ($q=.066$). These baseline contrasts quantify the beyond-baseline component of the joint evaluation.

\begin{table*}[htbp]
\centering
\small
\setlength{\tabcolsep}{3.5pt}
\begin{tabular}{llrrrrrr}
\toprule
Model & Stock & Task non-HOLD & Overconf. non-HOLD & Uplift & Task align. & Herding align. & Uplift \\
\midrule
Qwen3-8B & AAPL & 3.5 & 100.0 & 96.5 & 3.6 & 64.4 & 60.8 \\
Qwen3-8B & TSLA & 6.1 & 100.0 & 93.9 & 5.9 & 72.7 & 66.7 \\
Qwen3-14B & AAPL & 24.1 & 99.8 & 75.7 & 16.6 & 72.3 & 55.6 \\
Qwen3-14B & TSLA & 41.6 & 100.0 & 58.4 & 32.7 & 85.5 & 52.9 \\
GPT-5.4-mini & AAPL & 10.0 & 61.4 & 51.4 & 9.1 & 63.8 & 54.7 \\
GPT-5.4-mini & TSLA & 24.9 & 74.9 & 50.0 & 22.2 & 78.0 & 55.8 \\
Gemini-2.5-Flash & AAPL & 58.0 & 100.0 & 42.0 & 38.4 & 82.6 & 44.2 \\
Gemini-2.5-Flash & TSLA & 77.5 & 100.0 & 22.5 & 59.0 & 91.3 & 32.3 \\
\bottomrule
\end{tabular}
\caption{Zero-Shot directional checks. Activity uses 102 recorded decisions; alignment uses 101 price transitions and means \textsc{buy} after a rise or \textsc{sell} after a fall. Values are percentages; uplift is in percentage points.}
\label{tab:directional_zs}
\end{table*}

For the Loss-Aversion condition, we operationalize the intended feedback response as increased \textsc{hold} after negative rather than positive previous-action returns:
$\Delta H=P(\mathrm{HOLD}_t\mid r_{t-1}<0)-P(\mathrm{HOLD}_t\mid r_{t-1}>0)$.
Table~\ref{tab:loss_feedback} reports the baseline-relative contrast $\Delta\Delta H$, defined as Loss-Aversion $\Delta H$ minus matched no-persona $\Delta H$. Zero-return feedback is excluded, and paired estimates require both loss and gain observations in each rollout.

\begin{table*}[htbp]
\centering
\small
\begin{tabular}{llrrrr}
\toprule
Model & Stock & Paired $K$ & $\Delta\Delta H$ & Sign-flip $p$ & BH $q$ \\
\midrule
Qwen3-8B & AAPL & 5 & -.619 & .0625 & .100 \\
Qwen3-8B & TSLA & 4 & .331 & .3750 & .429 \\
Qwen3-14B & AAPL & 5 & -.223 & .0625 & .100 \\
Qwen3-14B & TSLA & 5 & -.277 & .0625 & .100 \\
GPT-5.4-mini & AAPL & 5 & -.000 & 1.0000 & 1.000 \\
GPT-5.4-mini & TSLA & 5 & .273 & .1250 & .167 \\
Gemini-2.5-Flash & AAPL & 5 & -.463 & .0625 & .100 \\
Gemini-2.5-Flash & TSLA & 5 & -.257 & .0625 & .100 \\
\bottomrule
\end{tabular}
\caption{Loss-Aversion feedback diagnostic relative to no-persona behavior.}
\label{tab:loss_feedback}
\end{table*}

\section{Complete Action-Diversity and Uncertainty Results}
\label{appendix:visualizations}

Table~\ref{tab:entropy_full} reports the complete entropy results omitted from the main text. Values use all 102 recorded actions and are macro-averaged across AAPL and TSLA.

\begin{table*}[htbp]
\centering
\small
\begin{tabular}{llrrr}
\toprule
Model & Method & Overconfidence & Loss Aversion & Herding \\
\midrule
GPT-5.4-mini & Zero-Shot & 1.03 & .59 & 1.03 \\
 & One-Shot & .81 & .56 & .89 \\
 & CoT & .85 & .54 & .94 \\
Gemini-2.5-Flash & Zero-Shot & .01 & .71 & .78 \\
 & One-Shot & .00 & .73 & .86 \\
 & CoT & .01 & .73 & .77 \\
Qwen3-8B & Zero-Shot & .26 & .45 & .77 \\
 & One-Shot & .11 & .36 & .67 \\
 & CoT & .32 & .49 & .79 \\
Qwen3-14B & Zero-Shot & .06 & .25 & .83 \\
 & One-Shot & .02 & .61 & .80 \\
 & CoT & .17 & .33 & .81 \\
\bottomrule
\end{tabular}
\caption{Stock-aware macro-averaged action entropy.}
\label{tab:entropy_full}
\end{table*}

\begin{figure}[htbp]
  \centering
  \includegraphics[width=\columnwidth]{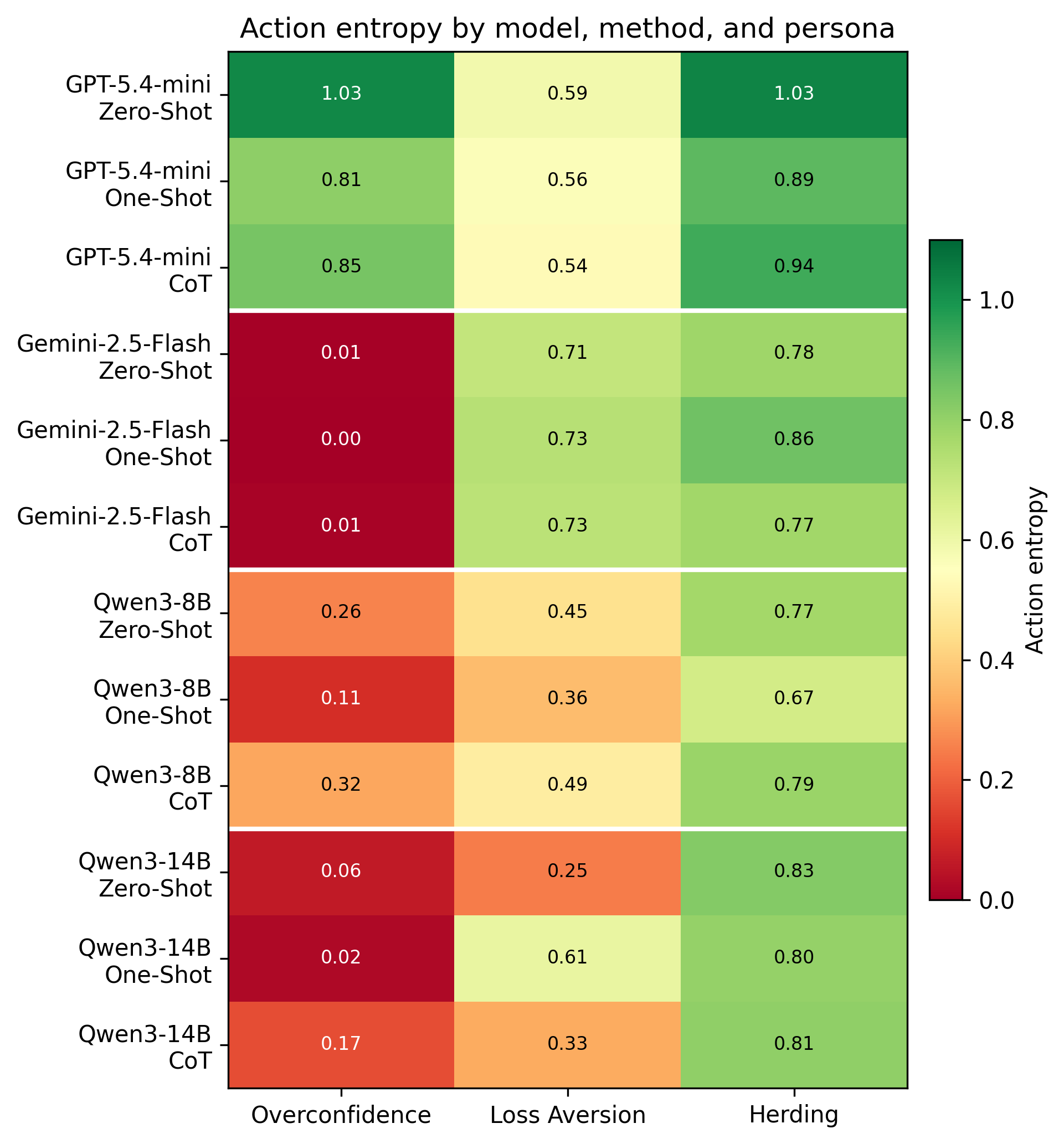}
  \caption{Action entropy across models, injection methods, and personas.}
  \label{fig:entropy_heatmap}
\end{figure}

Entropy is not a monotonic proxy for stability. Across the 20 prespecified model--method--stock conditions used in the targeted analysis, its Spearman correlation with within-persona return distance is $\rho=.032$ (permutation $p=.902$; bootstrap 95\% confidence interval, CI, $[-.446,.489]$). Its correlation with return B/W is $\rho=-.716$ ($p=.0008$; CI $[-.934,-.293]$). These results motivate joint interpretation with action separation and state sensitivity.

Table~\ref{tab:main_return_ci} reports 95\% Student's $t$ intervals for mean final cumulative log return in all main conditions ($K=5$). For example, GPT-5.4-mini/Overconfidence/Zero-Shot mean final-return intervals are $[-31.96,6.30]\%$ on AAPL and $[-31.60,22.90]\%$ on TSLA; their SDs are 15.40\% and 21.95\%. In contrast, Gemini-2.5-Flash/Overconfidence/Zero-Shot on TSLA has identical final returns across runs (SD 0).

\begin{table*}[htbp]
\centering
\footnotesize
\setlength{\tabcolsep}{3pt}
\begin{adjustbox}{max width=\textwidth}
\begin{tabular}{lllrrr}
\toprule
Model & Method & Stock & Overconfidence & Loss Aversion & Herding \\
\midrule
\multirow{6}{*}{GPT-5.4-mini}
& \multirow{2}{*}{Zero-Shot} & AAPL & $-12.83\;[-31.96, 6.30]$ & $4.52\;[-6.69, 15.73]$ & $-3.07\;[-10.95, 4.81]$ \\
& & TSLA & $-4.35\;[-31.60, 22.90]$ & $22.79\;[14.04, 31.54]$ & $21.48\;[6.67, 36.29]$ \\
\cmidrule(lr){2-6}
& \multirow{2}{*}{One-Shot} & AAPL & $-39.91\;[-52.73, -27.08]$ & $0.09\;[-6.78, 6.96]$ & $-14.28\;[-20.95, -7.60]$ \\
& & TSLA & $15.91\;[-12.78, 44.60]$ & $14.55\;[-10.80, 39.91]$ & $20.89\;[11.42, 30.37]$ \\
\cmidrule(lr){2-6}
& \multirow{2}{*}{CoT} & AAPL & $-29.10\;[-36.73, -21.46]$ & $16.76\;[9.02, 24.49]$ & $-1.38\;[-13.60, 10.84]$ \\
& & TSLA & $2.74\;[-14.41, 19.90]$ & $-4.10\;[-25.41, 17.22]$ & $-8.80\;[-20.65, 3.04]$ \\
\midrule
\multirow{6}{*}{Gemini-2.5-Flash}
& \multirow{2}{*}{Zero-Shot} & AAPL & $-19.01\;[-19.42, -18.61]$ & $28.19\;[18.79, 37.59]$ & $-1.01\;[-10.43, 8.41]$ \\
& & TSLA & $-9.05\;[-9.05, -9.05]$ & $1.37\;[-31.24, 33.98]$ & $28.37\;[20.51, 36.24]$ \\
\cmidrule(lr){2-6}
& \multirow{2}{*}{One-Shot} & AAPL & $-19.16\;[-19.16, -19.16]$ & $10.74\;[-1.32, 22.81]$ & $-2.85\;[-8.35, 2.65]$ \\
& & TSLA & $-9.05\;[-9.05, -9.05]$ & $21.13\;[0.79, 41.46]$ & $33.20\;[27.45, 38.95]$ \\
\cmidrule(lr){2-6}
& \multirow{2}{*}{CoT} & AAPL & $-19.33\;[-20.38, -18.28]$ & $22.44\;[-0.11, 44.98]$ & $-9.15\;[-13.98, -4.32]$ \\
& & TSLA & $-9.05\;[-9.05, -9.05]$ & $21.77\;[-31.77, 75.31]$ & $24.52\;[13.02, 36.03]$ \\
\midrule
\multirow{6}{*}{Qwen3-8B}
& \multirow{2}{*}{Zero-Shot} & AAPL & $-8.02\;[-8.51, -7.53]$ & $-2.64\;[-7.22, 1.93]$ & $-6.93\;[-15.11, 1.26]$ \\
& & TSLA & $25.04\;[22.60, 27.49]$ & $9.30\;[4.48, 14.12]$ & $100.26\;[96.02, 104.50]$ \\
\cmidrule(lr){2-6}
& \multirow{2}{*}{One-Shot} & AAPL & $-3.51\;[-17.95, 10.94]$ & $15.10\;[8.41, 21.78]$ & $-24.75\;[-28.62, -20.89]$ \\
& & TSLA & $-3.41\;[-22.49, 15.67]$ & $6.48\;[-2.20, 15.16]$ & $33.73\;[8.29, 59.18]$ \\
\cmidrule(lr){2-6}
& \multirow{2}{*}{CoT} & AAPL & $-2.76\;[-9.09, 3.56]$ & $1.01\;[-4.71, 6.72]$ & $-15.15\;[-34.11, 3.82]$ \\
& & TSLA & $48.37\;[17.66, 79.08]$ & $13.12\;[11.32, 14.92]$ & $43.10\;[35.35, 50.86]$ \\
\midrule
\multirow{6}{*}{Qwen3-14B}
& \multirow{2}{*}{Zero-Shot} & AAPL & $-25.20\;[-29.70, -20.71]$ & $19.34\;[11.04, 27.64]$ & $-1.06\;[-8.51, 6.39]$ \\
& & TSLA & $-2.80\;[-14.09, 8.50]$ & $8.33\;[2.70, 13.96]$ & $27.51\;[14.98, 40.05]$ \\
\cmidrule(lr){2-6}
& \multirow{2}{*}{One-Shot} & AAPL & $-20.64\;[-23.40, -17.88]$ & $4.60\;[-9.74, 18.94]$ & $-2.56\;[-13.31, 8.19]$ \\
& & TSLA & $-8.95\;[-15.83, -2.07]$ & $-5.32\;[-28.82, 18.18]$ & $6.09\;[2.17, 10.00]$ \\
\cmidrule(lr){2-6}
& \multirow{2}{*}{CoT} & AAPL & $-23.15\;[-35.04, -11.25]$ & $16.03\;[9.91, 22.16]$ & $-6.81\;[-11.78, -1.84]$ \\
& & TSLA & $3.37\;[-12.34, 19.08]$ & $8.57\;[-0.05, 17.20]$ & $25.41\;[14.02, 36.80]$ \\
\bottomrule
\end{tabular}
\end{adjustbox}
\caption{Mean final cumulative log return (\%) with 95\% Student's $t$ intervals over five rollouts.}
\label{tab:main_return_ci}
\end{table*}

\begin{figure*}[htbp]
  \centering
  \includegraphics[width=\textwidth]{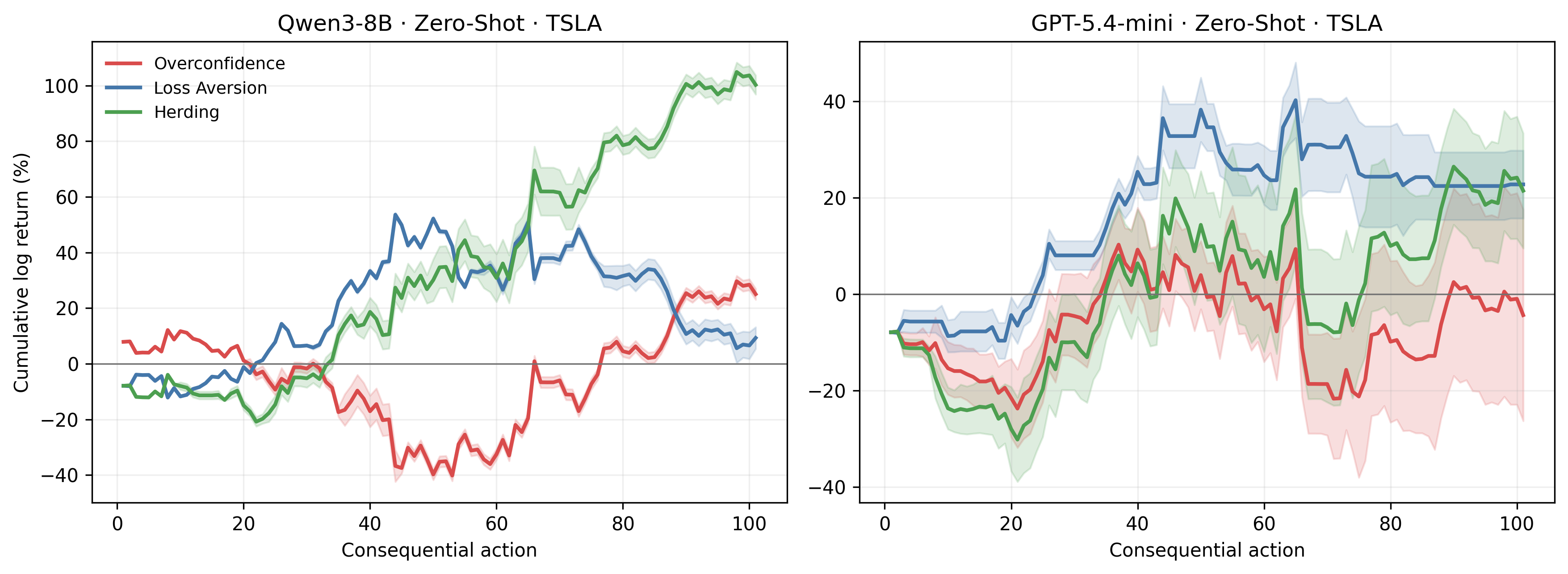}
  \caption{Zero-Shot cumulative-return trajectories on TSLA (mean $\pm$ SD over five runs).}
  \label{fig:trajectory_comparison}
\end{figure*}

\section{One-Shot Scenario Analyses}
\label{appendix:one_shot}

The mixed-evidence scenario defined in Appendix~\ref{appendix:persona_prompts} is the primary One-Shot condition. It contains 120 trajectories: four models, two stocks, three personas, and five runs. All primary return- and action-separation tests survive BH correction within their respective test families.

\begin{table*}[htbp]
\centering
\small
\begin{tabular}{llrrrr}
\toprule
Model & Stock & Return B/W & Return $q$ & Action B/W & Action $q$ \\
\midrule
GPT-5.4-mini & AAPL & 3.401 & .000114 & 6.111 & .000114 \\
GPT-5.4-mini & TSLA & 1.476 & .000200 & 4.098 & .000114 \\
Gemini-2.5-Flash & AAPL & 3.760 & .000114 & 11.147 & .000114 \\
Gemini-2.5-Flash & TSLA & 6.745 & .000114 & 6.422 & .000114 \\
Qwen3-8B & AAPL & 4.026 & .000114 & 26.479 & .000114 \\
Qwen3-8B & TSLA & 2.703 & .000114 & 14.881 & .000114 \\
Qwen3-14B & AAPL & 2.145 & .000200 & 4.951 & .000114 \\
Qwen3-14B & TSLA & 3.119 & .000114 & 4.702 & .000114 \\
\bottomrule
\end{tabular}
\caption{Primary One-Shot separation under the shared mixed-evidence demonstration.}
\label{tab:oneshot_primary}
\end{table*}

Against the no-persona baseline, all 24 action contrasts and 19 of 24 return contrasts are significant after correction. The five nonsignificant return contrasts are all Loss Aversion. Overconfidence non-HOLD uplift ranges from 22.6 to 96.4 percentage points; Herding alignment uplift ranges from 28.1 to 84.2 points; Loss-Aversion HOLD after a supplied loss ranges from 39.9\% to 100\%.

\subsection{Demonstration-Scenario Ablations}

We use two ablations. The positive-evidence scenario, denoted Scenario P, presents strong results, a 1.80\% price rise, bullish commentary, and no negative development; all three demonstrations select \texttt{BUY}. The negative-evidence scenario, denoted Scenario N, presents weak results, a 2.40\% fall, bearish commentary, and no positive development. Its Overconfidence and Herding demonstrations select \texttt{SELL}, while Loss Aversion selects \texttt{HOLD}. All use the same system-message architecture and JSON schema as the primary scenario.

The exact Scenario P rationales are: Overconfidence, ``The operating results and price response both confirm a strong opportunity. I trust this reading and will act decisively with a one-period positive exposure.''; Loss Aversion, ``The strong results and broad confirmation reduce near-term downside risk. I will take a measured one-period positive exposure while remaining alert to a reversal.''; Herding, ``The price rise, strong volume, bullish analyst commentary, and positive public sentiment all point in the same direction. I will follow the clear consensus with a one-period positive exposure.''

The exact Scenario N rationales are: Overconfidence, ``The earnings miss, weaker guidance, and price breakdown form a clear downside signal. I trust this reading and will act decisively with a one-period negative exposure.''; Loss Aversion, ``The outlook is negative, but taking a new directional position after a sharp move could add avoidable loss. I will stay neutral for one period to protect capital.''; Herding, ``The falling price, bearish analyst commentary, and negative public sentiment show a clear market consensus. I will follow the crowd with a one-period negative exposure.''

\begin{table*}[htbp]
\centering
\footnotesize
\setlength{\tabcolsep}{3pt}
\begin{tabular}{lllrrrr}
\toprule
Model & Scenario & Stock & Return B/W & Return $q$ & Action B/W & Action $q$ \\
\midrule
GPT-5.4-mini & Positive & AAPL & 2.264 & .000133 & 3.989 & .000133 \\
 & Positive & TSLA & 2.055 & .000133 & 3.515 & .000133 \\
 & Negative & AAPL & 1.212 & .035410 & 4.242 & .000133 \\
 & Negative & TSLA & 2.099 & .000686 & 2.916 & .000133 \\
Gemini-2.5-Flash & Positive & AAPL & 2.623 & .000474 & 5.155 & .000133 \\
 & Positive & TSLA & 4.413 & .000133 & 6.715 & .000133 \\
 & Negative & AAPL & 1.870 & .000133 & 6.113 & .000133 \\
 & Negative & TSLA & 2.833 & .000246 & 5.043 & .000133 \\
Qwen3-8B & Positive & AAPL & 2.021 & .000133 & 6.261 & .000133 \\
 & Positive & TSLA & 2.479 & .000133 & 7.718 & .000133 \\
 & Negative & AAPL & 3.864 & .000246 & 15.148 & .000133 \\
 & Negative & TSLA & 1.330 & .002538 & 7.055 & .000133 \\
Qwen3-14B & Positive & AAPL & 2.665 & .000133 & 3.207 & .000133 \\
 & Positive & TSLA & 2.449 & .000133 & 4.180 & .000133 \\
 & Negative & AAPL & 1.093 & .121388 & 2.400 & .000133 \\
 & Negative & TSLA & 1.165 & .038706 & 2.313 & .000133 \\
\bottomrule
\end{tabular}
\caption{One-Shot demonstration-scenario ablations. The positive scenario uses \texttt{BUY} for every persona.}
\label{tab:oneshot_scenarios}
\end{table*}

Thirty-one of 32 exploratory tests are significant; the sole exception is Qwen3-14B/AAPL return separation in the negative scenario. Persona separation therefore persists when the demonstrated action is identical across personas, although the demonstration still influences behavior.

\begin{table}[htbp]
\centering
\small
\resizebox{\columnwidth}{!}{%
\begin{tabular}{lrrr}
\toprule
Scenario & Overconfidence & Loss Aversion & Herding \\
\midrule
Mixed evidence & 65.5--100.0 & 50.9--88.5 & 42.0--55.1 \\
Positive evidence & 67.4--92.7 & 0.4--5.2 & 28.5--40.8 \\
Negative evidence & 15.2--41.3 & 63.5--89.3 & 33.5--50.0 \\
\bottomrule
\end{tabular}
}
\caption{Demonstration-action match-rate ranges across models (percent).}
\label{tab:oneshot_copying}
\end{table}

Matched-run comparisons show that changing the demonstration modifies 23.5\%, 20.1\%, and 26.6\% of consequential actions on average for mixed--positive, mixed--negative, and positive--negative comparisons, respectively; corresponding return-curve distances are .104, .112, and .112. Scenario content matters, but does not erase within-scenario persona separation.

\section{Cross-Sector Zero-Shot Replication}
\label{appendix:bmrn}

We add BioMarin Pharmaceutical (BMRN), a biotechnology equity, as a Zero-Shot replication under the same 2025-01-02 to 2025-05-30 window, action semantics, prompts, four models, and five runs. The matrix contains 80 trajectories: four models times one no-persona and three persona conditions times five runs.

\begin{table*}[htbp]
\centering
\footnotesize
\setlength{\tabcolsep}{3pt}
\begin{tabular}{llrrrrr}
\toprule
Model & Condition & Final CR (mean $\pm$ SD) & BUY & HOLD & SELL & Entropy \\
\midrule
Qwen3-8B & No-persona & $-1.05\pm0.11$ & 1.8 & 98.2 & 0.0 & .089 \\
 & Overconfidence & $-14.68\pm2.29$ & 99.4 & 0.6 & 0.0 & .031 \\
 & Loss Aversion & $4.67\pm1.58$ & 0.0 & 95.8 & 4.2 & .170 \\
 & Herding & $-14.05\pm5.40$ & 33.9 & 63.0 & 3.2 & .730 \\
Qwen3-14B & No-persona & $.96\pm2.84$ & 5.3 & 94.5 & 0.2 & .217 \\
 & Overconfidence & $-14.75\pm4.45$ & 93.9 & 5.9 & 0.2 & .235 \\
 & Loss Aversion & $13.77\pm7.08$ & 1.4 & 22.2 & 76.4 & .590 \\
 & Herding & $7.61\pm3.61$ & 51.5 & 3.2 & 45.3 & .806 \\
GPT-5.4-mini & No-persona & $.37\pm.60$ & 4.6 & 95.4 & 0.0 & .185 \\
 & Overconfidence & $-29.41\pm5.92$ & 54.7 & 45.1 & 0.2 & .688 \\
 & Loss Aversion & $4.62\pm6.61$ & 0.8 & 91.9 & 7.3 & .304 \\
 & Herding & $6.83\pm8.53$ & 43.4 & 25.0 & 31.7 & 1.072 \\
Gemini-2.5-Flash & No-persona & $-2.85\pm8.22$ & 8.5 & 91.1 & 0.4 & .313 \\
 & Overconfidence & $-13.54\pm.00$ & 100.0 & 0.0 & 0.0 & .000 \\
 & Loss Aversion & $17.73\pm9.35$ & 1.8 & 81.4 & 16.8 & .538 \\
 & Herding & $9.40\pm2.95$ & 52.3 & 3.0 & 44.8 & .803 \\
\bottomrule
\end{tabular}
\caption{BMRN condition results. Returns and action rates are percentages; entropy uses the 101 consequential actions.}
\label{tab:bmrn_conditions}
\end{table*}

All 12 persona--baseline return tests and all 12 action tests have exact $p=q=.007937$. Return/action B/W ranges are 2.699--6.291 and 3.658--15.345, respectively.

\begin{figure*}[htbp]
  \centering
  \includegraphics[width=\textwidth]{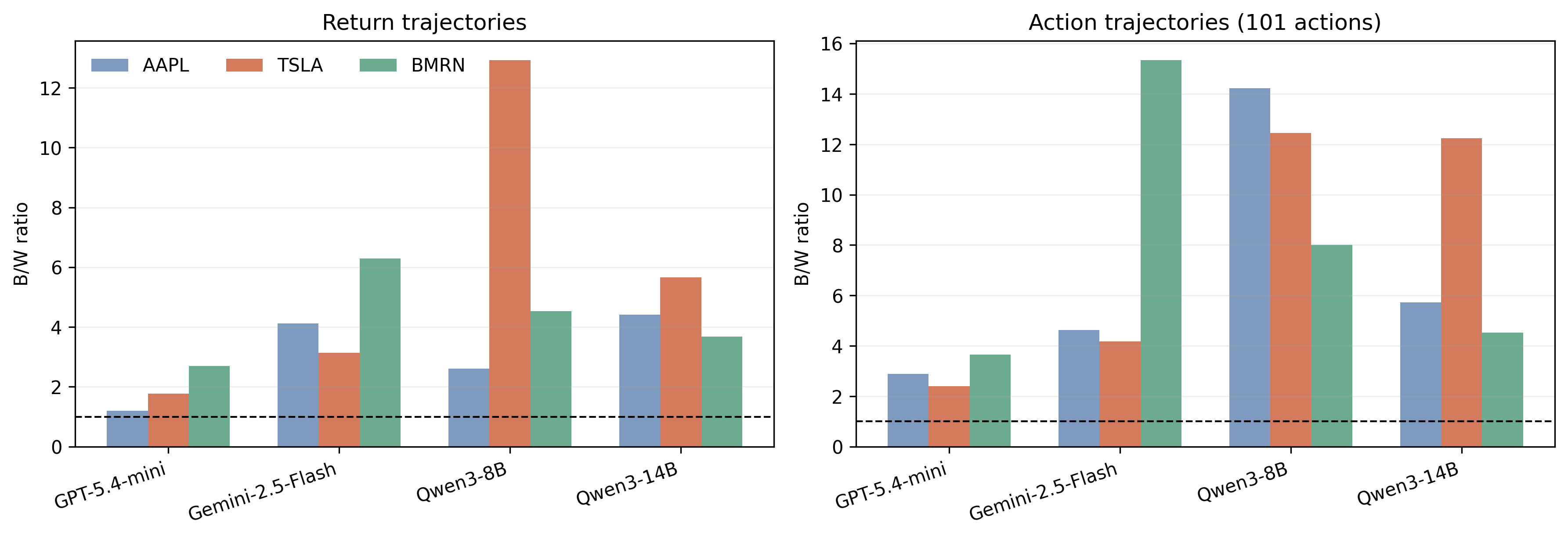}
  \caption{Zero-Shot persona separation across AAPL, TSLA, and BMRN. The dashed line marks B/W$=1$.}
  \label{fig:cross_asset_bw}
\end{figure*}

\begin{table*}[htbp]
\centering
\small
\setlength{\tabcolsep}{4pt}
\begin{tabular}{lrrrrrr}
\toprule
& \multicolumn{3}{c}{Overconfidence non-HOLD uplift} & \multicolumn{3}{c}{Herding alignment uplift} \\
\cmidrule(lr){2-4}\cmidrule(lr){5-7}
Model & AAPL & TSLA & BMRN & AAPL & TSLA & BMRN \\
\midrule
Qwen3-8B & .964 & .939 & .976 & .608 & .667 & .267 \\
Qwen3-14B & .754 & .586 & .885 & .556 & .529 & .935 \\
GPT-5.4-mini & .515 & .499 & .503 & .547 & .558 & .709 \\
Gemini-2.5-Flash & .414 & .226 & .911 & .442 & .323 & .913 \\
\bottomrule
\end{tabular}
\caption{Baseline-relative directional effects across assets.}
\label{tab:cross_asset_direction}
\end{table*}

Every model--asset value in Table~\ref{tab:cross_asset_direction} is positive. This extends the directional pattern to a biotechnology equity, but the shared date window does not test multiple market regimes.

\section{Second Non-Overlapping AAPL Stream}
\label{appendix:aapl_r2}

\paragraph{Design and market context.}
We evaluate a second AAPL stream, denoted R2, from 2 June to 24 October 2025. It contains 102 NYSE decision sessions and 101 consequential actions per rollout. Four models each complete five seeded rollouts under no-persona, Overconfidence, Loss Aversion, and Herding, yielding 80 trajectories. The no-persona condition removes persona text. Persona descriptions, Zero-Shot prompting, generation settings, and action/return semantics follow the shared protocol in Appendix~\ref{appendix:protocol}. Non-trading-day evidence is rolled forward to the next session; neither momentum nor future-price fields enter the prompts. All 80 trajectories pass the output audit, with zero repair calls and zero fallback actions.

The original AAPL window (2 January--30 May 2025) has a realized buy-and-hold return of $-17.4\%$ and maximum drawdown of $-30.2\%$; R2 has corresponding values of $+30.5\%$ and $-5.6\%$. These price-path statistics characterize the contrasting market contexts. The two windows also differ in textual-evidence construction, so cross-window differences are descriptive rather than estimates of a market-regime effect.

\paragraph{Trajectory separation.}
Table~\ref{tab:aapl_r2_bw} reports persona-only B/W over the three persona conditions. All eight ratios exceed one and are significant under 10,000 global label permutations. Table~\ref{tab:aapl_r2_exact} reports all 12 persona--baseline contrasts. Each five-versus-five exact test enumerates all $\binom{10}{5}=252$ labelings; BH correction is applied separately to the 12 return and 12 action tests. Every contrast has $p=q=.00794$. Return distances use the 101 realized cumulative-log-return points, and action distances use normalized Hamming distance over the 101 consequential actions.

\paragraph{Temporal coverage.}
The entire R2 stream postdates the April 2025 Qwen3 release and Gemini-2.5-Flash's reported January 2025 knowledge cutoff. All nine persona--baseline contrasts involving these three models remain significant in both action and return space ($q=.00794$; Table~\ref{tab:aapl_r2_exact}). Persona-conditioned separation therefore recurs on a market stream later than these temporal references.

\begin{table}[htbp]
\centering
\small
\setlength{\tabcolsep}{3pt}
\begin{adjustbox}{max width=\columnwidth}
\begin{tabular}{lrrrr}
\toprule
& \multicolumn{2}{c}{Return} & \multicolumn{2}{c}{Action} \\
\cmidrule(lr){2-3}\cmidrule(lr){4-5}
Model & B/W & $p$ & B/W & $p$ \\
\midrule
Qwen3-8B & 4.979 & $<.00010$ & 13.741 & $<.00010$ \\
Qwen3-14B & 4.886 & $<.00010$ & 4.664 & .00020 \\
GPT-5.4-mini & 3.251 & $<.00010$ & 4.938 & $<.00010$ \\
Gemini-2.5-Flash & 9.690 & $<.00010$ & 7.221 & $<.00010$ \\
\bottomrule
\end{tabular}
\end{adjustbox}
\caption{AAPL R2 Zero-Shot persona separation. Action distances use 101 consequential actions.}
\label{tab:aapl_r2_bw}
\end{table}

\begin{table*}[htbp]
\centering
\small
\setlength{\tabcolsep}{4pt}
\begin{tabular}{llrrrr}
\toprule
Model & Persona & Return $\Delta D$ & Return $q$ & Action $\Delta D$ & Action $q$ \\
\midrule
Qwen3-8B & Overconfidence & .0974 & .00794 & .9901 & .00794 \\
 & Loss Aversion & .0254 & .00794 & .0327 & .00794 \\
 & Herding & .0354 & .00794 & .8139 & .00794 \\
Qwen3-14B & Overconfidence & .0946 & .00794 & .7817 & .00794 \\
 & Loss Aversion & .0386 & .00794 & .6209 & .00794 \\
 & Herding & .0141 & .00794 & .8492 & .00794 \\
GPT-5.4-mini & Overconfidence & .0555 & .00794 & .6535 & .00794 \\
 & Loss Aversion & .0185 & .00794 & .0661 & .00794 \\
 & Herding & .0237 & .00794 & .7061 & .00794 \\
Gemini-2.5-Flash & Overconfidence & .0922 & .00794 & .7515 & .00794 \\
 & Loss Aversion & .0612 & .00794 & .1987 & .00794 \\
 & Herding & .0282 & .00794 & .7733 & .00794 \\
\bottomrule
\end{tabular}
\caption{AAPL R2 exact persona--baseline tests. $\Delta D$ discounts within-condition dispersion as defined in Appendix~\ref{appendix:baseline}; BH correction is separate for the 12 return and 12 action tests.}
\label{tab:aapl_r2_exact}
\end{table*}

\paragraph{Behavioral interpretation.}
Overconfidence produces non-\textsc{hold} rates of 83.0--100.0\%, with Qwen3-8B and Gemini selecting \textsc{buy} at every step. Herding alignment is .610/.935/.820/.950 for Qwen3-8B, Qwen3-14B, GPT, and Gemini, respectively, compared with .010/.034/.055/.135 for their no-persona baselines. These directions recur on the rising AAPL stream. Loss Aversion remains model-dependent: Qwen3-8B and GPT select \textsc{hold} on 95.0\% and 93.3\% of consequential actions, whereas Qwen3-14B selects \textsc{sell} on 77.4\%. These proportions characterize the action preferences induced by the Loss-Aversion prompt.

Four model--condition cells produce an identical action sequence across all five seeds: Qwen3-8B no-persona, Qwen3-8B Overconfidence, Gemini Overconfidence, and Gemini Herding. The two Overconfidence cells are all-\textsc{buy}; Gemini Herding instead follows a changing action sequence aligned with price direction. Thus, identical repeated trajectories and single-action collapse are distinct diagnostics. B/W establishes separation, while action diversity and directional checks determine how that separation arises. Together, the R2 results extend persona-conditioned separation to a second AAPL price path and reproduce the paper's distinction between separation and coherent, context-sensitive persona policies.

\section{Persona-Paraphrase Robustness}
\label{appendix:paraphrase}

This analysis compares the original descriptions with the two semantically equivalent variants in Appendix~\ref{appendix:persona_prompts}. It combines 120 original and 240 new persona trajectories. Let $W$ be repeated-run distance under the same persona and wording, $P$ be distance for the same persona under different wordings, and $B$ be distance for different personas under the same wording. We quantify paraphrase robustness by requiring $B>P$ and a paired-bootstrap 95\% interval for $B-P$ above zero.

\begin{table*}[htbp]
\centering
\small
\setlength{\tabcolsep}{3pt}
\begin{tabular}{lllrrrrr}
\toprule
Model & Stock & Metric & $W$ & $P$ & $B$ & $B/P$ & 95\% CI for $B-P$ \\
\midrule
GPT-5.4-mini & AAPL & Return & .0652 & .0687 & .0887 & 1.29 & [.0079,.0270] \\
 & AAPL & Action & .1631 & .2313 & .4771 & 2.06 & [.2261,.2628] \\
 & TSLA & Return & .1055 & .1223 & .1488 & 1.22 & [.0075,.0432] \\
 & TSLA & Action & .1793 & .2400 & .4278 & 1.78 & [.1750,.1978] \\
Gemini-2.5-Flash & AAPL & Return & .0436 & .0757 & .1811 & 2.39 & [.0910,.1162] \\
 & AAPL & Action & .1122 & .2163 & .6673 & 3.09 & [.4439,.4602] \\
 & TSLA & Return & .0695 & .0906 & .3288 & 3.63 & [.2108,.2765] \\
 & TSLA & Action & .0963 & .1609 & .6868 & 4.27 & [.5185,.5343] \\
Qwen3-8B & AAPL & Return & .0446 & .0786 & .0774 & .99 & [-.0048,.0058] \\
 & AAPL & Action & .0892 & .3241 & .5804 & 1.79 & [.2440,.2751] \\
 & TSLA & Return & .0560 & .2048 & .2944 & 1.44 & [.0746,.1044] \\
 & TSLA & Action & .0878 & .2958 & .5925 & 2.00 & [.2862,.3087] \\
Qwen3-14B & AAPL & Return & .0458 & .0539 & .1593 & 2.95 & [.0833,.1234] \\
 & AAPL & Action & .1602 & .2063 & .7134 & 3.46 & [.4873,.5273] \\
 & TSLA & Return & .0897 & .1050 & .3118 & 2.97 & [.1764,.2435] \\
 & TSLA & Action & .0967 & .1237 & .6805 & 5.50 & [.5415,.5759] \\
\bottomrule
\end{tabular}
\caption{Persona separation ($B$) relative to wording variation ($P$) and repeated-run variation ($W$).}
\label{tab:paraphrase_wpb}
\end{table*}

Fifteen of 16 rows meet the criterion, including all eight action rows. The exception is Qwen3-8B/AAPL return. Separately, persona separation is significant within 24/24 wording-specific action comparisons and 23/24 return comparisons. Same-persona wording tests detect differences in 20/24 action and 11/24 return comparisons.

\begin{figure*}[htbp]
  \centering
  \includegraphics[width=.82\textwidth]{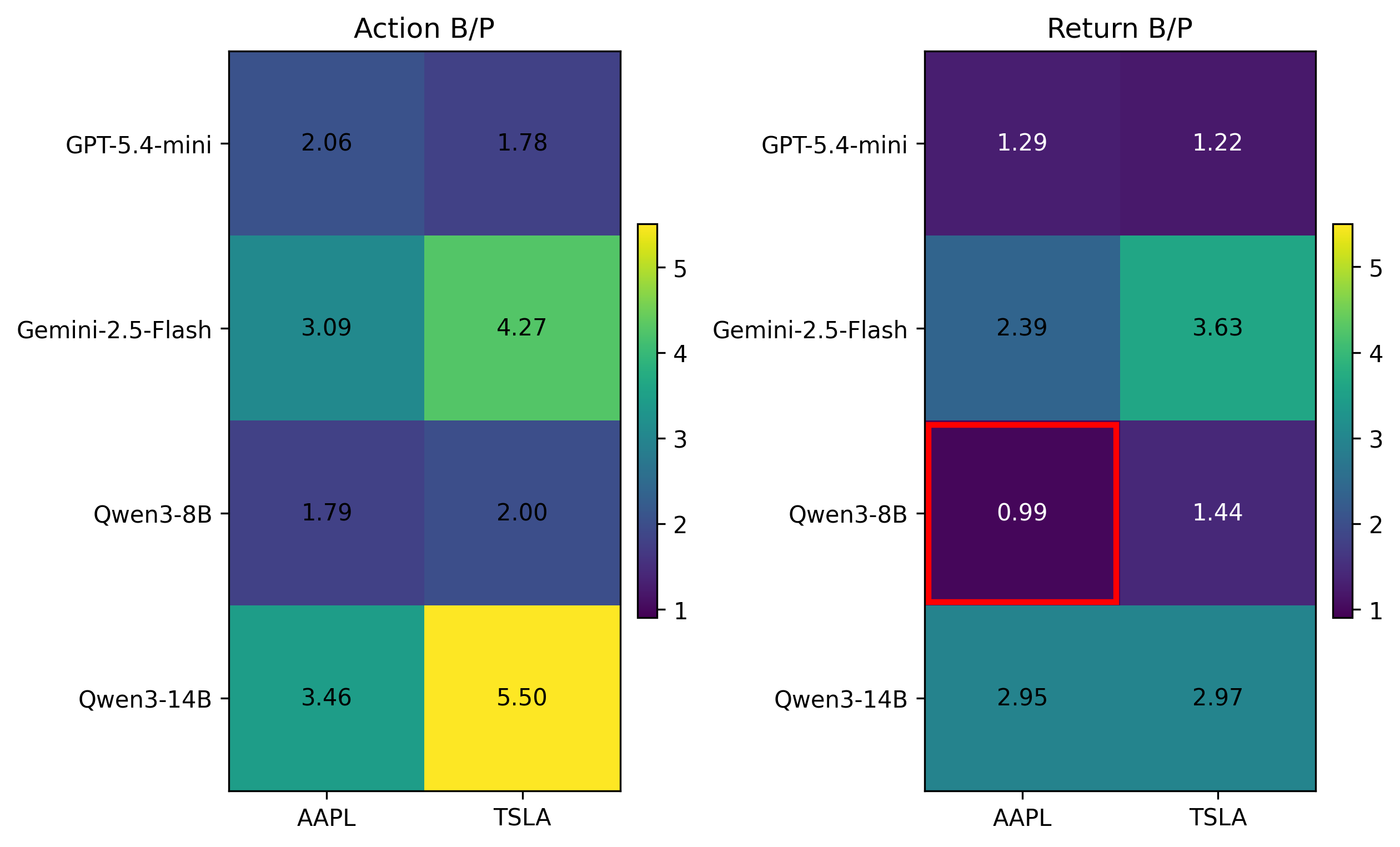}
  \caption{Ratio of persona separation to paraphrase variation. The outlined cell is the sole $B/P<1$ result.}
  \label{fig:paraphrase_bp}
\end{figure*}

\section{Fine-Tuning Ablation Details}
\label{appendix:sft_details}

We fully fine-tune Qwen3-8B on 2,202 training samples with 246 validation samples for five epochs. Training uses two GPUs, per-device batch size 1, gradient accumulation 16, global effective batch approximately 32, learning rate $10^{-5}$ with cosine scheduling and 0.1 warmup, AdamW with weight decay .01, and bfloat16 (BF16) arithmetic. Checkpoint 100 (epoch 1.45) has the lowest observed validation loss, .032996, and is used for evaluation. Full SFT uses LLaMA-Factory with the Qwen3 chat template, a 4,096-token sequence-length cutoff, gradient checkpointing, and DeepSpeed ZeRO-3 with CPU optimizer offload. Validation and checkpoint saving occur every 20 steps.

Labels are supplied by GPT-4.1-mini using financial sentiment and momentum features. Teacher and deterministic rule labels differ for 785/2,448 samples (32.1\%). In the label set, Flat, Long, and Short map to \textsc{hold}, \textsc{buy}, and \textsc{sell}. The Loss-Aversion teacher distribution contains 813 Flat, 3 Long, and 0 Short labels; its all-\textsc{hold} result is therefore interpreted as an outcome of this supervision design. The GPT-4.1-mini labeling prompts include sentiment features from \texttt{ProsusAI/finbert}~\cite{araci2019finbert}.
Positive, neutral, and negative probabilities are averaged over non-overlapping 510-token chunks of the news and filing texts; net sentiment is positive minus negative.

\begin{table}[htbp]
\centering
\small
\resizebox{\columnwidth}{!}{%
\begin{tabular}{lrrrrr}
\toprule
& \multicolumn{2}{c}{Return B/W} & \multicolumn{3}{c}{Entropy} \\
\cmidrule(lr){2-3}\cmidrule(lr){4-6}
Method & AAPL & TSLA & Overconf. & Loss Av. & Herding \\
\midrule
Zero-Shot & 2.61 & 12.93 & .26 & .45 & .77 \\
One-Shot & 4.03 & 2.70 & .11 & .36 & .67 \\
CoT & 2.25 & 2.60 & .32 & .49 & .79 \\
SFT & 4.44 & 3.40 & .58 & .00 & .86 \\
\bottomrule
\end{tabular}}
\caption{Qwen3-8B SFT ablation and prompt-based conditions.}
\label{tab:sft_vs_prompt}
\end{table}

SFT changes action preferences but does not consistently remove instability or collapse. Its Loss-Aversion entropy of zero is consistent with the highly imbalanced supervision reported above.

\section{Case Studies}
\label{appendix:case_study}

\subsection{Stereotyped Policy Reduction}

Gemini-2.5-Flash under Overconfidence and Zero-Shot on TSLA selects \textsc{buy} at all 102 steps in every run. Entropy and switch rate are zero, \textsc{buy} follows both rises and falls 100\% of the time, and every run ends at $-9.05\%$. Thus, zero return SD reflects state-insensitive action collapse.

\begin{table}[htbp]
\centering
\small
\begin{tabular}{lrrr}
\toprule
Date & Price & Action & Cumulative return \\
\midrule
2025-01-02 & 379.28 & \textsc{buy} & 0.00\% \\
2025-01-03 & 410.44 & \textsc{buy} & 7.90\% \\
2025-01-06 & 411.05 & \textsc{buy} & 8.04\% \\
2025-01-07 & 394.36 & \textsc{buy} & 3.90\% \\
2025-01-08 & 394.94 & \textsc{buy} & 4.05\% \\
2025-01-10 & 394.74 & \textsc{buy} & 4.00\% \\
2025-01-13 & 403.31 & \textsc{buy} & 6.14\% \\
2025-01-14 & 396.36 & \textsc{buy} & 4.40\% \\
2025-01-15 & 428.22 & \textsc{buy} & 12.14\% \\
2025-01-16 & 413.82 & \textsc{buy} & 8.72\% \\
\multicolumn{4}{c}{$\cdots$ all intermediate actions are \textsc{buy} $\cdots$} \\
2025-05-29 & 358.43 & \textsc{buy} & $-5.65\%$ \\
2025-05-30 & 346.46 & \textsc{buy} & $-9.05\%$ \\
\bottomrule
\end{tabular}
\caption{Representative all-\textsc{buy} trajectory on TSLA.}
\label{tab:gemini_overconf_tsla}
\end{table}

\subsection{Noisy Policy Switching}

GPT-5.4-mini under Overconfidence and Zero-Shot on AAPL produces final returns from $-27.26\%$ to $11.73\%$ across five identical-condition runs, a 39.00-point spread. The mean is $-12.83\%$ and the sample SD is 15.40\%, illustrating high-diversity but unstable behavior.

\begin{figure}[htbp]
  \centering
  \includegraphics[width=\columnwidth]{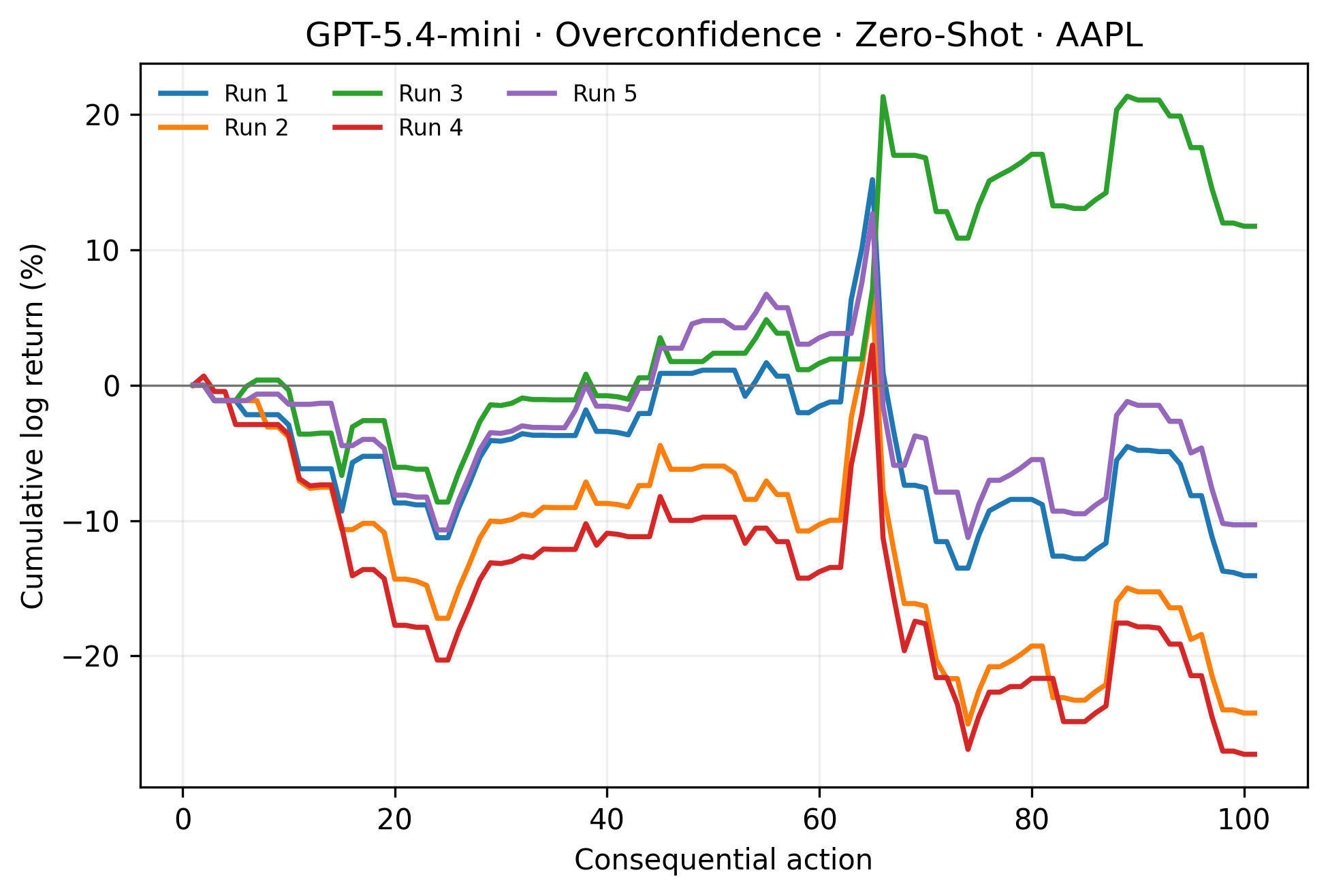}
  \caption{Five GPT-5.4-mini Overconfidence Zero-Shot trajectories on AAPL.}
  \label{fig:gpt_overconf_diverge}
\end{figure}

\end{document}